\documentclass[pdflatex,sn-mathphys-num]{sn-jnl}
\usepackage{graphicx} 
\usepackage{xcolor}
\usepackage{amsmath}
\usepackage{amssymb}
\usepackage{hyperref}
\usepackage{cleveref}
\usepackage{fullpage}
\usepackage{algorithm}
\usepackage{algpseudocode}
\usepackage{siunitx}
\usepackage{orcidlink}

\usepackage{multirow}%
\usepackage{amsmath,amssymb,amsfonts}%
\usepackage{amsthm}%
\usepackage{mathrsfs}%
\usepackage[title]{appendix}%
\usepackage{xcolor}%
\usepackage{textcomp}%
\usepackage{manyfoot}%
\usepackage{booktabs}%
\usepackage{algorithmicx}%
\usepackage{listings}%

\DeclareMathOperator*{\argmin}{arg\,min}

\newcommand{\E}{\mathbb{E}}
\newcommand{\x}{\boldsymbol{x}}
\newcommand{\y}{\boldsymbol{y}}

\newcommand{\data}{\boldsymbol{d}}
\newcommand{\w}{\boldsymbol{w}}
\newcommand{\z}{\boldsymbol{z}}

\begin{document}

\title[Article Title]{Neural Optimal Design of Experiment for Inverse Problems}

\author*[1]{\fnm{John E.} \sur{Darges\orcidlink{0009-0007-9059-8921}}}\email{jdarges@emory.edu}

\author[2]{\fnm{Babak Maboudi} \sur{Afkham\orcidlink{0000-0003-3203-8874}}}\email{babak.maboudi@oulu.fi}
\equalcont{These authors contributed equally to this work.}

\author[1]{\fnm{Matthias} \sur{Chung\orcidlink{0000-0001-7822-4539}}}\email{matthias.chung@emory.edu}
\equalcont{These authors contributed equally to this work.}

\affil*[1]{\orgdiv{Department of Mathematics}, \orgname{Emory University}, \orgaddress{ \city{Atlanta, Georgia 30322}, \country{United States}}}

\affil[2]{\orgdiv{Unit of Mathematical Sciences}, \orgname{University of Oulu}, \orgaddress{\city{Oulu}, \state{Pentti Kaiteran katu 1, 90570} \country{Finland}}}

\abstract{We introduce \emph{Neural Optimal Design of Experiments} (NODE), a learning-based framework for optimal experimental design in inverse problems that avoids classical bi-level optimization and indirect sparsity regularization. NODE jointly trains a neural reconstruction model and a fixed-budget set of continuous design variables—representing sensor locations, sampling times, or measurement angles—within a single optimization loop. By optimizing measurement locations directly rather than weighting a dense grid of candidates, the proposed approach enforces sparsity by design, eliminates the need for $\ell^1$ tuning, and substantially reduces computational complexity. We validate NODE on an analytically tractable exponential-growth benchmark, on \texttt{MNIST} image sampling, and illustrate its effectiveness on a real-world sparse-view X-ray CT example. In all cases, NODE outperforms baseline approaches, demonstrating improved reconstruction accuracy and task-specific performance.}


\maketitle

\section{Introduction}

The quality and reliability of experimental outcomes are fundamentally linked to how experiments are designed. Optimal Experimental Design (OED) provides a mathematical framework for selecting measurement locations, times, or sensing modalities so as to maximize the information gained about quantities of interest, subject to constraints on cost, noise, and available resources. At its core, OED navigates a central trade-off: extracting the richest possible information from observations while respecting practical and computational limitations.

OED has found widespread use across scientific and engineering disciplines. In pharmaceutical research, it informs the design of clinical trials that balance information gain against cost and risk \cite{dmitrienko2007pharmaceutical,beg2019application,chung2012experimental}; in agriculture and industrial experimentation, it guides efficient field trials and process optimization \cite{walsh2022overview,zou2016application}; in the social sciences and marketing, it improves the reliability of surveys and preference studies \cite{murnane2010methods,kuhfeld1994efficient}; and in environmental and health studies, it supports data-driven impact assessment and monitoring \cite{goos2011optimal}. Across these settings, common design principles include efficiency, robustness to noise and model uncertainty, flexibility, and cost-effectiveness.

From a mathematical perspective, OED problems are typically formulated through optimality criteria derived from statistical estimation theory. A classical approach relies on the Fisher information matrix, which quantifies the sensitivity of observables to unknown parameters. Popular criteria include D-optimality, which maximizes the determinant of the Fisher information; A-optimality, which minimizes the average variance of parameter estimates; and E-optimality, which maximizes the smallest eigenvalue, ensuring uniform precision across parameters \cite{pukelsheim2006optimal,atkinson2007optimum}. While these criteria provide a principled basis for experimental design, translating them into scalable algorithms becomes increasingly challenging in complex inverse problems.

Inverse problems aim to infer parameters or states from indirect and noisy observations and are often ill-posed: small perturbations in data can induce large changes in reconstructions. In such settings—common in medical imaging, geophysics, and dynamical systems—the placement and timing of measurements play a decisive role in stabilizing inference and mitigating noise amplification. Consequently, OED is not merely beneficial but often essential for obtaining meaningful solutions.

Two major paradigms dominate OED for inverse problems. The Bayesian formulation incorporates prior information and typically seeks designs that maximize expected information gain between prior and posterior distributions \cite{chaloner1995bayesian,alexanderian2020oedreview}. While uncertainty-aware and theoretically appealing, these objectives often require expensive sampling, nested quadrature, or variational approximations, limiting their scalability in high-dimensional or nonlinear settings. An alternative is bi-level optimization, in which an upper-level design problem is coupled to a lower-level regularized inverse problem \cite{haber2008numerical,horesh2010optimal,chung2012experimental,ruthotto2018optimal}. This approach directly links design quality to reconstruction performance, while reducing the computational cost compared to the Bayesian approaches it remains computationally demanding: each design update typically requires solving a full inverse problem, often with forward and adjoint PDE solves, and nonconvexity in the inner problem can lead to nondifferentiability in the outer objective.

A common practical challenge in both paradigms is the enforcement of sparsity. Measurement budgets are frequently controlled indirectly through $\ell^1$-based regularization or relaxation strategies, requiring careful tuning of penalty parameters and often yielding designs whose effective number of measurements is difficult to predict or control. These issues motivate the search for alternative formulations that couple design and inversion more tightly while avoiding nested optimization loops and delicate regularization trade-offs.

A recent and influential step in this direction is deep optimal experimental design \cite{siddiqui2024deep}, which reframes OED as a single-level learning problem. In this approach, a neural estimator is trained jointly with design variables embedded directly into the network inputs, using simulated data across noise levels. By learning the inversion map and the experimental design simultaneously, the method avoids inner inverse solves and demonstrates strong performance on ODE-based models \cite{siddiqui2024deep}. This perspective is conceptually appealing and highlights the potential of learning-based OED to overcome classical computational bottlenecks.

However, this single-level approach begin from a dense discretized grid of candidate measurements and promote sparsity through shrinkage operators or combinatorial masks. As a result, computational complexity remains tied to the initial discretization, and sparsity is still controlled indirectly through regularization heuristics. While the bi-level bottleneck is removed, the challenge of managing sparse designs persists.

\paragraph{Contributions.}
Motivated by these observations, we propose a new framework, termed \emph{Neural Optimal Design of Experiments (NODE)}, that retains the advantages of single-level joint training while enforcing sparsity directly by design. Rather than starting from a dense candidate grid, we assume a fixed measurement budget from the outset and treat the corresponding measurement locations or times as continuous variables. These design variables are optimized jointly with a neural reconstruction model, allowing measurement locations to move freely within the admissible domain. In this way, sparsity is not induced through penalization but prescribed explicitly by the experimental budget, eliminating the need for $\ell^1$ tuning and decoupling computational complexity from an artificial discretization.

Building on this fixed-budget formulation, we further introduce an adaptive and iterative extension of NODE for sequential experimental settings. Previously collected measurements are incorporated into the design process, enabling subsequent measurement locations to be optimized conditionally on observed data. This adaptivity aligns naturally with practical experimental workflows and enhances robustness in the presence of noise or model mismatch.

\paragraph{Structure.}
The remainder of the paper is organized as follows. Section~\ref{sec:background} briefly reviews classical and modern OED approaches with an emphasis on inverse problems. Section~\ref{sec:method} presents the NODE framework, including the continuous fixed-budget formulation and its adaptive extension. In Section~\ref{sec:ode}, we validate the approach on a benchmark ODE problem with analytically characterized optimal designs. Section~\ref{sec:numerics} reports numerical experiments on \texttt{MNIST} image sampling (Section~\ref{sec:mnist}) and sparse-view CT reconstruction (Section~\ref{sec:ct}). We conclude in Section~\ref{sec:conclusion} with a discussion of limitations and directions for future work.

\section{Background}\label{sec:background}
We consider inverse problems~\cite{tarantola2005,hansen2010} governed by a forward model \begin{equation}\label{eq:forward-model}
    \y = f(\z,\x).
\end{equation}
Here, $\z$ represents the domain inputs of the model. For example, $\z$ may be a time variable or spatial coordinates. These inputs can lie on a continuous or discrete domain space. The output is the vector $\y$. The model is determined by parameters $\x$, so that $f(\cdot,\x)$ maps inputs to outputs. For some fixed parameters $\x$, $f(\cdot,\x)$ results in a collection of points $\big(\z,\y\big)$. For example, the points form a trajectory when $f$ is the solution to a system of ordinary differential equations, a surface when $f$ is the solution to a system of partial differential equations, or an image when $f$ maps pixel locations to grayscale values. 

When we observe real-world phenomena that can be described by a mathematical model in the form of~\eqref{eq:forward-model}, we usually only observe at a finite number of locations in the input domain
$\{(\z_i,\y_i)\}_{i=1}^N$. An observation operator $P$ selects which locations out of the whole that we observe. Solving an inverse problem means recovering the parameters $\x$ that, through the model, yielded the observations. Measurement-taking adds noise to the data, so that what we truly observe is 
$$
    \data=\y+\varepsilon,
$$ 
where $\varepsilon\sim\mathcal{M}$ is independent across different measurements. The noise distribution is commonly assumed to be a Gaussian distribution centered at zero. Inverse problems are notoriously ill-posed due to this presence of noise and because the unknown parameters often outnumber the available data. The Bayesian framework~\cite{stuart2010inverse} is a popular approach to inverse problems, as a way of propagating uncertainty to the solution and regularizing the problem. The solution is a probability distribution informed by prior knowledge on the parameters and the observed data. This is the posterior distribution, as given by Bayes' rule
\begin{equation}
    \pi_{\mathrm{post}}(\x\mid\data) = \frac{\pi_{\text{pr}}(\x)\pi_\textrm{like}(\data\mid\x)}{\int \pi_{\text{pr}}(\x)\pi_\textrm{like}(\data\mid\x)\, \text{d} \x}.
\end{equation}
The prior distribution $\pi_{\text{pr}}$ summarizes our knowledge and assumptions about the parameters before we observe the experimental data. The likelihood $\pi_\mathrm{like}$ is an unnormalized distribution for the observation data that is conditioned on the parameters. It describes, the likelihood, given some fixed values for the parameters, of observing the observations under our assumptions for the noise. 

\subsection{Optimal Experimental Design}
How often and where should we measure so that the inverse problem yields accurate and stable solutions?  This question lies at the heart of \emph{optimal experimental design} (OED) for inverse problems.
In OED, design parameters $\w\subset \mathcal W$ dictate where measurements are taken through the observation operator $P_{\w}$ which operates on $\y$. We observe the noisy data point $\data=P_{\w}(\y)+\varepsilon$. The admissible set $\mathcal{W}$ encodes possible physical or geometric constraints, restricting the design to a feasible region. Solving an OED problem means determining which choice of design parameters will most likely give us an inverse problem which we can solve accurately. Fundamentally, an OED problem is
\begin{equation}
\min_{\w} \quad \E_{\x,\varepsilon}\operatorname{Loss}_{\text{design}}(\widehat{\x}(\w,\varepsilon), \x)  ,\quad \x\sim\pi_{\text{pr}},\,\varepsilon\sim\mathcal{M},
\end{equation}
where $\operatorname{Loss}_{\text{design}}$ measures how well our solution, using the design $\w$, $\widehat{\x}(\w,\varepsilon)$ reconstructs the true parameters $\x$ on average. 

Early work by Haber and 
collaborators~\cite{haber2008numerical} and Horesh et al.~\cite{horesh2010optimal} established the \emph{bi-level optimization problem} formulation of OED and proposed efficient computational strategies in the context of linear and nonlinear inverse problems, respectively. In this setting, the outer problem selects the design $\w$, while the inner problem solves the corresponding inverse problem for reconstructing the unknown parameters $\x$, with the goal of ensuring stable and accurate recovery. 
A generic bi-level OED formulation can be stated as a Bayesian risk minimization problem,
\begin{gather}
    \min_{\w} \quad \mathbb{E}_{\x,\varepsilon}  \ \
    \operatorname{Loss}_{\text{design}}(\widehat{\x}(\w,\varepsilon), \x)  + \mathcal{R}(\w),
    \label{eq:bi-level-OED-outer} \\[6pt]
    \widehat{\x}(\w,\varepsilon) = \argmin_{\tilde \x} \ \ \operatorname{Loss}_{\text{inv}}\!\left(P_{\w}(\y(\tilde\x),\, P_{\w}(\y(\x))+\varepsilon\right) + \mathcal{D}(\tilde\x).
    \label{eq:bi-level-OED-inner}
\end{gather}
Here, the design parameters $\w$ represent nonnegative weights associated with each candidate location or time where a sensor can record a measurement. A larger weight reflects that observations should be made at the corresponding location. A regularization term $\mathcal{R}(\w)$ is added to the design loss.
The reconstruction $\widehat{\x}(\w,\varepsilon)$ is obtained by solving the inner inverse problem~\eqref{eq:bi-level-OED-inner}, where the data misfit loss $\operatorname{Loss}_{\text{inv}}$ reflects the assumed statistical noise model and is complemented by a regularization term $\mathcal{D}(\x)$ encoding prior assumptions on the parameter $\x$. 

The choice of the design loss $\operatorname{Loss}_{\text{design}}$ in~\eqref{eq:bi-level-OED-outer} reflects the criterion under which the experimental design is optimized. Among others, there are three classical choices--A-optimality, D-optimality, and E-optimality~\cite{pukelsheim2006optimal}. A-optimality is minimizing the average variance in each parameter estimate and can translate both to minimizing the mean squared error (MSE) between the average parameter estimate (either the posterior mean or the maximum a posterior point) and true parameter. This is equivalent to minimizing the trace of the inverse of the information matrix, which corresponds to a Gaussian covariance matrix when the posterior distribution is Gaussian~\cite{aoptalen}. D-optimality is an information-based criterion which prioritizes maximizing the distance, measured via Kullback-Liebler divergence, between the posterior and prior distributions. This is often formulated as maximizing the determinant of the information matrix (or minimizing the log-determinant of the inverse of the information matrix), \cite{doptalen}. E-optimality minimizes the worst possible variance in one of the parameter estimates by maximizing the smallest eigenvalue of the information matrix (or by minimizing the largest eigenvalue of the inverse of the information matrix). These are the most common approaches when the aim is accurate parameter estimation. When the parameter estimates are just a mean towards estimating some other quantity of interest or prediction, the array of optimality criteria becomes numerous, and OED becomes goal-oriented~\cite{Attia2018,neuberger-quadratic,madhavan2025}.

In practice, the expectation in~\eqref{eq:bi-level-OED-outer} is approximated by a sample average over draws $\{\x_k\}_{k=1}^K$ from $\pi_{\text{pr}}$ and noise realizations. This turns the outer objective into an empirical risk. Synthetic training data can be generated when lacking sufficient experimental data, but this assumes that true experimental data are consistent with the same underlying distribution.  Each evaluation of the risk requires solving the inner inverse problems, which are handled with standard optimization algorithms (e.g., Gauss--Newton) and warm-started across outer iterations. 

A practical way to solve \Cref{eq:bi-level-OED-outer} is to begin with an \emph{overcomplete} catalogue of $M$ admissible design actions---for instance, a grid of candidate sensor locations or measurement times. A binary selection vector $\w \in \{0,1\}^M$ may encode which actions are active, where $w^{(j)}=1$ indicates that the $j$-th candidate is chosen. Since this constitutes an NP-hard combinatorial optimization, the problem is typically relaxed by allowing continuous weights $\w \in [0,1]^M$. The need to approximate discrete solutions motivates one to promote sparsity by augmenting the objective with an $\ell^1$ penalty 
\[
\mathcal{R}(\w) = \lambda \|\w\|_1, \qquad \lambda > 0,
\]
where $\lambda$ is a design (sparsity) regularization hyperparameter. Theoretical guarantees from compressed sensing establish that (e.g., under Restricted Isometry Property (RIP) conditions) that $\ell^1$ relaxations are highly effective in promoting sparsity \cite{candes2005decoding}. In the OED setting, small values of $\lambda$ correspond to weak budgetary constraints and yield denser designs, whereas large values of $\lambda$ enforce stronger sparsity and select only a few informative measurement locations. A Pareto trade-off between reconstruction quality and design sparsity determines the selected ``optimal'' design. For the outer updates, gradients with respect to $\w$ can be obtained efficiently by differentiating through the inner solver, using adjoint-based techniques \cite{haber2008numerical,horesh2010optimal}. The $\ell^1$ penalty is typically handled with proximal or continuation strategies that gradually promote sparsity, and after convergence, small weights are pruned or thresholded to yield a discrete subset of informative measurement points.

While conceptually appealing, the bi-level OED formulation is notoriously challenging to solve in practice. Every evaluation of the outer objective requires the solution of a full inverse problem, often repeated for multiple parameter draws in order to approximate the expectation. 
Many inverse problems, especially ones with high-dimensional parameters, are challenging to solve once, let alone multiple times. The feasibility of solving large-scale problems generally relies assumptions that the prior and noise distributions are Gaussian. The presence of nonlinearities in such problems merits using approximations, especially a Gaussian approximation of the posterior~\cite{alexanderian2024optimal,chowdhary2025robust}, but also the quadratic approximation of the goal function in~\cite{neuberger-quadratic}. These approximation methods are necessary, but introduce bias. Ill-posed settings further complicate this by the need for careful selection of priors and regularization parameters, which can destabilize the optimization and strongly influence the resulting design. Moreover, sparsity enforcement through $\ell^1$ penalties only indirectly controls the number of sensors: as the penalty weight increases, many design variables are indeed driven to zero, but the remaining non-zero parameters may shrink artificially, leading to ambiguous or physically unrealistic designs.  Thus, although the bi-level framework directly links experimental decisions to reconstruction accuracy, its computational cost and sensitivity to modeling choices strongly motivate the search for alternatives.

Well-posedness of the outer objective in~\eqref{eq:bi-level-OED-outer} requires properties such as convexity, uniqueness, and differentiability of the inner solution map in~\eqref{eq:bi-level-OED-inner}. These assumptions are rarely satisfied in realistic inverse problems, making the outer function discontinuous and its gradients unstable~\cite{ruthotto2018optimal}.  Stabilization of the inner problem through strong priors may reduce this sensitivity but also bias reconstructions, limit the effectiveness of the design, and introduce hyperparameter dependencies. While Bayesian global optimization approaches have been proposed to mitigate such effects~\cite{chung2021efficient}, they remain computationally prohibitive in high-dimensional settings. In addition to these challenges, practical OED is further complicated by issues such as nonconvex outer objectives, difficulties in discretizing relaxed designs, and sensitivity to model or prior misspecification.

\subsection{Single-level OED with Neural Networks}

To overcome some of the major challenges associated with bi-level optimization approaches to OED, Siddiqui, Rahmim, and Haber proposed \emph{deep optimal experimental design} in 2024 \cite{siddiqui2024deep}. The central idea is to bypass the nested bi-level structure by training a neural network $\Phi_\theta$, with parameters $\theta$, that simultaneously learns both the inversion map and the experimental design. Within this single-level framework, the forward model is used only to generate synthetic training data across a distribution of parameters and noise realizations. The network is then trained end-to-end to minimize reconstruction error while treating the design variables as trainable quantities embedded in the input layer. In this way, inversion and design are learned jointly, avoiding the repeated forward–inverse loops characteristic of classical bi-level formulations.

Rather than solving an inner inverse problem for each design $\w$, a neural network is used to approximate the maximum a posteriori (MAP) estimator directly,
\begin{equation}
    \Phi_\theta\big(\y(\w;\x,\varepsilon), \w\big)
    \approx \arg\max_{\x} \ \pi_{\text{post}}\big(\x \mid \y(\w;\x,\varepsilon)\big).
    \label{eq:neuralnetwork}
\end{equation}
Here, parameters $\x \sim \pi_{\text{pr}}$ and noise $\varepsilon \sim \mathcal{N}(\boldsymbol{0},\sigma^2 \mathbf{I})$ are sampled to generate synthetic observations via the forward model~\eqref{eq:forward-model} at a design specified by $\w$. The network outputs an estimate of $\x$ directly from the data, with the design variables $\w$ and noise implicitly embedded in the inputs. In practice, $\w$ re-weights the input data element-wise via a Hadamard product prior to the first network layer. As a result, the estimator $\widehat{\x}=\Phi_\theta(\cdot)$ is obtained through a single forward evaluation rather than by solving an optimization problem.

The estimator and design are trained jointly by minimizing a Bayes risk that combines reconstruction accuracy with data consistency,
\begin{equation}
    \min_{\theta,\w \in \mathcal{W}} \ 
    \mathbb{E}_{\x, \varepsilon}
    \|\widehat\x - \x\|_2^2
    + \gamma \|\y(\w,\widehat\x) - \data(\w; \x, \varepsilon)\|_2^2 
    + \lambda \|\w\|_1, 
    \quad \widehat\x=\Phi_\theta\big(\y(\w;\x,\varepsilon), \w\big),
    \label{eq:deep-oed}
\end{equation}
where $\gamma,\lambda \ge 0$. This objective can be interpreted as a variant of A-optimality, with the $\ell^1$ penalty promoting sparsity in the design. The second term enforces data consistency, ensuring that the reconstructed parameters reproduce the observed data, and may be viewed as a physics-informed regularization analogous to those used in PINNs \cite{raissi2019physics}.

The principal advantage of this framework is that it eliminates the nested structure of classical bi-level OED: the forward model is used solely to generate training data, while both the estimator parameters $\theta$ and the design variables $\w$ are optimized jointly within a single learning problem. This significantly reduces computational cost and avoids repeated inverse solves.

Despite these advantages, a computational challenge remains when designs are defined over a large set of candidate measurement locations. For PDE-based inverse problems in particular, fine spatial or temporal discretizations are often required, leading to high-dimensional design vectors $\w$ and correspondingly large input tensors $\y(\w)$. This tight coupling between design resolution and network input dimensionality substantially increases model complexity and training cost, motivating approaches that avoid reliance on dense, predefined candidate grids altogether.

To promote compact and informative designs within this setting, Siddiqui, Rahmim, and Haber explore several sparsity-inducing strategies for $\w$, including $\ell^1$ relaxations and discrete Tabu-search methods. This highlights that, even within a single-level formulation, achieving sparse and interpretable designs remains nontrivial and often requires additional algorithmic machinery — a difficulty that motivates the alternative, grid-free approach proposed in this work.

\section{Method}\label{sec:method}
Our approach is an evolution on deep OED of~\cite{siddiqui2024deep}, we refer to our method as NODE (neural optimal design of experiment). In our approach, a budget of $M$ sensors is specified and the design variables $\w$ represent the \textit{locations} of those $M$ sensors. This differs from traditional OED setups where $\w$ represent \textit{weights} on candidate locations. We follow the same philosophy in of leveraging neural networks for design optimization to avoid the complexity of bi-level formulations. This formulation overcomes two major limitations of the previous approach: it eliminates the need for tuning sparsity parameters and $\ell^1$ regularization, and it substantially reduces the dimensionality of the problem—designing over the number of sensors rather than the number of potential placement sites. To be more precise, let us consider a continuous and bounded design space $\mathcal{W}$, such as a time interval or a bounded spatial domain. Our objective is to select $M$ design points $\w=\{w^{(m)}\}_{m = 1}^M \subset \mathcal{W}$ that minimize a given loss function. 

To train our estimator we go hand in hand with the objective laid out in \Cref{eq:deep-oed} and the design jointly, the empirical risk that blends a risk of the inversion process with a data-consistency risk is considered, i.e., it simplifies to
\begin{equation}
    \min_{\theta, \w \in \mathcal{W}} \ 
    \mathbb{E}_{\x, \varepsilon}
    \|\widehat{\x} - \x\|_2^2
        + \gamma \|P_{\w}\big(\y(\widehat\x)\big) - P_{\w}\big(\data(\x,\varepsilon)\big)\|_2^2, \quad \widehat{\x}=\Phi_\theta\big(P_{\w}(\y(\x)+\varepsilon\big),
    \qquad \gamma \geq 0.
    \label{eq:deep-oed2}
\end{equation}
While \Cref{eq:deep-oed} and \Cref{eq:deep-oed2} appear similar, the key distinction lies in the interpretation of the design variable $\w$. In the former, $\w$ represents weights associated with a predefined set of candidate data points, whereas in the latter, $\w$ directly parameterizes the measurement locations or observation times themselves. Consequently, the optimization is performed over the actual sensor positions or sampling times, rather than over a weighting of fixed candidates. This eliminates the need for sparsity-inducing regularization such as $\lambda \|\w\|_1$ as utilized in \Cref{eq:deep-oed}, while retaining an A-optimal design criterion.

Further, when the forward model is unavailable (e.g., as in the numerical experiment in \Cref{sec:mnist}), computationally prohibitive, or otherwise impractical to employ, we set $\gamma = 0$ and omit the corresponding term. The resulting formulation reduces to a purely data-driven learning objective, focused solely on recovering the unknown parameter $\x$ from the observed data.

Assuming~\eqref{eq:deep-oed2} is continuous and well-posed, we leverage universal approximation theorems to ensure that a neural network can approximate the mapping from design variables to the objective value. Consequently, when the problem is well-posed—meaning we have enough design points for the parameter complexity—we can establish theoretical bounds on the approximation error of the optimal design.
We note that these guarantees depend on the problem not being underdetermined. If the number of design points is insufficient, the theoretical approximation results do not apply. In summary, our method provides robust approximation guarantees under appropriate conditions.
Naturally, this shift from location-based to budget-based discretization requires approximating the underlying continuous placement in space (and time), but yields a far more scalable and interpretable design process, as we show in our numerical experiments.

\paragraph{Algorithm.} A pseudocode summary of the proposed approach is given in \Cref{alg:node}. Depending on whether an underlying mathematical forward model is available, the observations $y_k$ may either be computed on the fly via the forward operator and noise simulation or drawn from a precomputed training dataset.

\Cref{alg:node} jointly updates the design variables $\w$ and the parameters $\theta$ of the reconstruction network $\Phi_\theta$. At each iteration, a mini-batch of training samples is drawn and projected onto the current design via the projection operator $P_{\w}$, which extracts the measurements at the selected sensor locations. Observational noise is added when working with simulated or noise-free data to mimic realistic measurement conditions. The reconstruction network then maps the resulting measurements to an estimate of the unknown parameter $\x$, and both $\w$ and $\theta$ are updated jointly using gradient-based optimization.

While we explicitly separate the design variables $\w$ from the reconstruction network $\Phi_\theta$ for conceptual clarity and modularity, these components can alternatively be thought of a single neural architecture. Such a unified formulation is adopted, for example, in the deep optimal experimental design framework of~\cite{siddiqui2024deep}. The present separation, however, allows for greater modality in incorporating different design parameterizations and inversion models.

\begin{algorithm}
\caption{Neural optimal design of experiments (NODE)}\label{alg:node}
\begin{algorithmic}
\State \textbf{Inputs:} Training set $\{(\x_k,\y_k)\}_{k = 1}^K$, Architecture of $\Phi_\theta$, Noise distribution $\varepsilon\sim\mathcal{M}$, Sensor budget $M$, Design space $\mathcal{W}$
\State \textbf{Outputs:} Optimized locations $\w^*$, Reconstruction model $\Phi_{\theta^*}$
\State Initialize $\w_0$ \Comment{e.g., as the locations of highest variance across all $\y_k$} 
\State Initialize $\theta_0$
\For{$i=1,\dots,i_{\max}$}
    \State Sample batch $(\x_i,\y_i)$
    \State Project to current  design $\data_i=P_{\w_i}(\y_i)+\varepsilon$ 
    \Comment{Add observational noise if needed}
    \State Compute reconstruction $\hat{\x}=\Phi_{\theta_i}(\data_i)$
    \State Compute loss \Comment{Empirical version of \Cref{eq:deep-oed2}}
    \State Update $\w_i\to\w_{i+1},\theta_i\to\theta_{i+1}$
\EndFor
\State Set $\theta^*=\theta_{i_{\max}}$
\State Set $\w^*= \w_{i_{\max}}$ 
\end{algorithmic}
\end{algorithm}

\subsection{Interpolation over Finite Design Space}\label{sec:finte-node}
When the design space $\mathcal{W}$ is continuous, optimization of \Cref{eq:deep-oed2} is straightforward, as gradients with respect to the design variables $\w$ can be computed and standard gradient-based methods applied. In practice, however, observations may be restricted to finitely many admissible locations, so that $\mathcal{W}$ is a discrete subset of $M$ candidate points in the input domain. For a given parameter $\x$, the model output can then only be observed at these locations, yielding a finite vector $\y = [y^{(1)},\dots,y^{(M)}]^\top$.

To circumvent the resulting discrete optimization problem, we smoothly interpolate the observable outputs $\y$ over a continuously extended design space. We denote the interpolated outputs by $\tilde{\y}$ and the corresponding continuous design variables by $\tilde{\w}$. This continuous relaxation enables gradient-based optimization of the design variables within the learning loop. During training, the interpolated outputs $\tilde{\y}$ are projected onto $M$ sensor locations specified by $\tilde{\w}$, and the resulting measurements serve as input to the reconstruction network.

The discrete structure of the admissible design set can be exploited to construct an informed and inexpensive initialization of the sensor locations. Rather than initializing $\w$ at random, we select admissible locations that exhibit large empirical variance across the training dataset. Specifically, for each candidate location $w^{(m)}$, we compute the empirical variance
\begin{equation}\label{eq:init_variance}
\widehat{\mathrm{Var}}\bigl(w^{(m)}\bigr)
= \tfrac{1}{K-1}\sum_{k=1}^K\Bigl(y_k^{(m)}-\bar{y}^{(m)}\Bigr)^2,
\qquad \text{where } \quad
\bar{y}^{(m)} = \tfrac{1}{K}\sum_{k=1}^K y_k^{(m)}.
\end{equation}
The initial design $\widetilde{\w}_0$ is chosen as the $M$ locations with the largest variance values among all admissible candidates. This simple heuristic provides a reasonable proxy for informativeness and serves as a robust starting point for subsequent gradient-based optimization.

Upon completion of training, the optimized continuous design $\widetilde{\w}$ is projected back onto the admissible set $\mathcal{W}$. When $\mathcal{W}$ consists of integer-valued locations—as is the case for image-based designs defined on pixel grids—this projection can be performed by simple rounding $\widetilde{\w}$ to the nearest admissible locations, yielding the final design $\w^*$.

Several interpolation strategies may be employed to construct $\tilde{\y}$, including $n$-linear~\cite{getreuer2011linear}, $n$-cubic~\cite{cubic}, radial basis function~\cite{rbf}, or neural interpolation methods such as SIREN~\cite{siren}. While these approaches are often designed for high-fidelity resolution enhancement, such accuracy is not required for the present purpose. We therefore recommend linear interpolation, which is computationally inexpensive and sufficient for enabling effective gradient-based optimization of the design variables.

\subsection{Adaptive NODE} \label{sec:adaptive-NODE}

In many practical experimental settings, data are collected sequentially. For example, in medical X-ray computed tomography the measurement device rotates around the object and acquires a sequence of projections, while in active seismic imaging sources are deployed according to a prescribed acquisition sequence. In such scenarios, measurements collected at earlier stages can be leveraged to inform and optimize the selection of subsequent measurement locations. In this section, we introduce an \emph{adaptive NODE} framework that explicitly incorporates past measurements into the design of future experiments.

In a nutshell, adaptive NODE proceeds sequentially by fixing the previously learned $i-1$ design points and augmenting the NODE optimization by selecting an additional measurement location. At each step, the network is updated using all measurements acquired thus far, and a new design point is chosen to complement the existing design. Repeating this procedure yields an adaptive, data-informed experimental design.

Formally, suppose we are at the $i$-th step of the adaptive NODE procedure. We assume access to either a forward operator $f$ or an experimental device that has provided noisy measurements $\data^{i-1}$ corresponding to previously selected design parameters $\w^{i-1}$. To incorporate these past observations into \Cref{eq:deep-oed2}, we modify the reconstruction network to include the previously acquired data as a non-trainable input, yielding
\begin{equation} \label{eq:net-adaptive}
    \widehat{\x}
    = \Phi^{\mathrm{adapt}}_\theta\bigl(
    P_{\w}(\y(\x)+\varepsilon),
    P_{\w^{i-1}}(\y^{i-1})
    \bigr).
\end{equation}
Here, $\w    \subset \mathcal{W}$ denotes a proposed new set of design parameters to be optimized.

The corresponding adaptive NODE objective is then given by
\begin{equation} \label{eq:adaptive-cost}
    \min_{\theta,\, \w \in \mathcal{W}}
    \ \mathbb{E}_{\x,\varepsilon}
    \bigl\|\widehat{\x} - \x\bigr\|_2^2
    + \gamma
    \left\|
    \begin{bmatrix}
        P_{\w}\bigl(\y(\widehat{\x})\bigr) \\
        P_{\w^{i-1}}\bigl(\y^{i-1}\bigr)
    \end{bmatrix}
    -
    \begin{bmatrix}
        P_{\w}\bigl(\data(\x,\varepsilon)\bigr) \\
        P_{\w^{i-1}}\bigl(\data(\x,\varepsilon)\bigr)
    \end{bmatrix}
    \right\|_2^2,
    \qquad \gamma \ge 0.
\end{equation}
Solving this optimization problem yields a new set of design parameters $\w^\star$, which are then used with the forward operator $f$ or the experimental apparatus to obtain a new measurement realization $\y^\star$.

The design and observation sets are subsequently updated by concatenation: we define $\w^{i} = (\w^{i-1}, \w^\star)$ and $\y^{i} = (\y^{i-1}, \y^\star)$. This procedure is then repeated, enabling sequential, data-informed refinement of the experimental design.

Note that the input dimension of $\Phi^{\mathrm{adapt}}_{\theta}$ changes as new measurements are incorporated. Consequently, $\Phi^{\mathrm{adapt}}_{\theta}$ requires a dynamic architecture capable of accommodating an increasing number of inputs. One option is to assume a maximum input dimension and pad missing measurements with zeros; however, this can lead to an unnecessarily large memory footprint. Alternatively, as discussed in the previous section, the measurements $\y^{i}$ may be interpolated to a fixed-dimensional representation $\widetilde{\y}^{i}$, allowing the architecture of $\Phi^{\mathrm{adapt}}_{\theta}$ to remain fixed across all adaptive NODE iterations.

We emphasize that previously explored design parameters $\w^{i-1}$ are not excluded from the search space $\mathcal{W}$. In the presence of noise or unforeseen experimental artifacts—such as patient motion in medical imaging—newly acquired measurements may be insufficiently informative. In such cases, the adaptive NODE framework naturally allows for repeated measurements at previously selected locations until a desired level of information is obtained.

The adaptive NODE procedure continues until a specified stopping criterion is met, such as achieving a target reconstruction accuracy or exhausting a predefined measurement budget. The complete procedure is summarized in \Cref{alg:adaptive-node}.

\begin{algorithm}
\caption{Adaptive NODE}\label{alg:adaptive-node}
\begin{algorithmic}
\State \textbf{Inputs:} Training set $\{(\x_k,\y_k)\}_{k =1}^K$, Architecture of $\Phi_\theta$, Noise distribution $\varepsilon\sim\mathcal{M}$, Sensor incremental budget $M$, Design space $\mathcal{W}$, Initial design parameter $\w^0$ and measurement $\y^0$, and stopping rule $\mathcal L$.
\State \textbf{Outputs:} Optimized locations $\w^*$, Reconstruction model $\Phi_{\theta^*}$
\State Set $i=1$
\While{stopping rule $\mathcal L$ is not met}
\State Solve \eqref{eq:adaptive-cost} for $\theta^\star$ and $\w^\star$
\State Perform actual measurement corresponding to $\w^{\star}$ to obtain $\y^{\star}$
\State $[\w^\star,\w^{i}] \to \w^{i}$
\State $[\y^\star,\y^{i-1}] \to \y^{i}$
\State $i+1\to i$
\EndWhile
\end{algorithmic}
\end{algorithm}

\subsection{Exponential–Growth Benchmark}\label{sec:ode}

To rigorously assess our approach, we require a numerical experiment where the true optimal design is analytically known. Such a setting provides a clear benchmark: we can verify that our method achieves provably optimal accuracy rather than relying on empirical evidence alone. Kiefer’s seminal 1959 work \cite{kiefer1959optimum} offers broad structural principles for optimal experimental design, but these results remain general and do not resolve specific models. Following works \cite{fedorov1997model,dette2010optimal} provide details on optimal experimental design for regression models, however, here we focus on a very specific experiment for benchmarking.

We therefore develop a specific, fully analyzable benchmark problem that admits an exact characterization of the optimal design. We consider the exponential growth model 
\begin{equation}\label{eq:exp}
    y'(t) = \beta\, y(t) \quad \text{on } t\in[0,1] \quad  \text{with } y(0)=a.
\end{equation}
This model has been explored numerically \cite{siddiqui2024deep, chung2012experimental} and confirmed optimality only empirically (for a two point design only). Here we derive the optimal design explicitly and extend the result to any number of design points. Our analysis follows the spirit of Kiefer’s foundational theory while providing a closed-form solution for this specific nonlinear regression problem. We use this benchmark problem to rigorously test our approach, and it simultaneously offers an independent theoretical contribution.

For \Cref{eq:exp} with solution $y(t)=a\,\mathrm{e}^{\beta t}$, we consider multiplicative noise of the form
\[
\tilde y_j \;=\; y(t_j)\,\xi_j \;=\; a\,\mathrm{e}^{\beta t_j}\,\xi_j,
\qquad j=1,\dots,m,
\]
where $\{\xi_j\}$ are independent positive random variables with finite second logarithmic moment with $m\geq 2$.
We assume that
\[
\mathbb{E}\bigl(\log \xi_j\bigr)=0,
\qquad
\operatorname{Var}\bigl(\log \xi_j\bigr)=\sigma^2 < \infty,
\]
for all $j=1,\dots,m$, ensuring that the log-transformed noise is centered and admits a diagonal covariance structure with uniform variance across sampling times.
Define $y_j:=\log\tilde y_j$ and $\varepsilon_j:=\log\xi_j$, and collect $\mathbf{y}=[y_1,\ldots,y_m]^\top$ and $\boldsymbol{\varepsilon}=[\varepsilon_1,\ldots,\varepsilon_m]^\top$.  
Let $\mathbf{x}_{\text{true}}=[\alpha,\beta]^\top$ with $\alpha=\log a$, $\mathbf{t}=[t_1,\ldots,t_m]^\top$, and $\mathbf{A}(\mathbf{t})=[(1,t_1);\ldots;(1,t_m)]$.  
The log–linear model is $\mathbf{y}(\mathbf{t})=\mathbf{A}(\mathbf{t})\,\mathbf{x}_{\text{true}}+\boldsymbol{\varepsilon}$ with $\mathbb{E}(\boldsymbol{\varepsilon})=\mathbf{0}$ and $\operatorname{Cov}(\boldsymbol{\varepsilon})=\sigma^2\mathbf{I}$.
Note our zero-mean assumption entails no loss of generality for the design problem, since adding a deterministic shift to the log–observations does not affect the optimal placement of design points.

The \emph{inner parameter estimation problem} can now be formulated as follows:  
given $\mathbf{t}$, estimate $\mathbf{x}$ by least squares
\[
\hat{\mathbf{x}}(\mathbf{t})
= \arg\min_{\mathbf{x}}\|\mathbf{y}(\mathbf{t})-\mathbf{A}(\mathbf{t})\,\mathbf{x}\|_2^2,
\qquad
\mathbf{A}(\mathbf{t})^\top \mathbf{A}(\mathbf{t})\,\hat{\mathbf{x}}
= \mathbf{A}(\mathbf{t})^\top \mathbf{y}(\mathbf{t}).
\]
Assuming full \emph{column} rank of $\mathbf{A}(\mathbf{t})$ (i.e., the $t_j$ are not all identical), the solution is
\begin{equation}
\hat{\mathbf{x}}(\mathbf{t})
= \big(\mathbf{A}(\mathbf{t})^\top \mathbf{A}(\mathbf{t})\big)^{-1}
   \mathbf{A}(\mathbf{t})^\top \mathbf{y}(\mathbf{t})
= \mathbf{x}_{\text{true}}
  + \big(\mathbf{A}(\mathbf{t})^\top \mathbf{A}(\mathbf{t})\big)^{-1}
    \mathbf{A}(\mathbf{t})^\top \boldsymbol{\varepsilon}.
\label{eq:exp_inverse_solution}
\end{equation}

\medskip
This leads to the \emph{outer design problem}:  we seek design locations $\mathbf{t} \in [0,1]^m$ that minimize the
expected reconstruction error of given 
$\hat{\mathbf{x}}(\mathbf{t})$, i.e.,
\[
\min_{\mathbf{t}\in[0,1]^m}  \ F(\mathbf{t}) = \mathbb{E}\ \|\hat{\mathbf{x}}(\mathbf{t})-\mathbf{x}_{\text{true}}\|_2^2.
\]
Using \Cref{eq:exp_inverse_solution} and the fact that
$\mathbb{E}(\boldsymbol{\varepsilon}\boldsymbol{\varepsilon}^\top)=\sigma^2\mathbf{I}$, we obtain
\begin{align*}
\mathbb{E}\,\|\hat{\mathbf{x}}(\mathbf{t})-\mathbf{x}_{\text{true}}\|_2^2
&= \sigma^2 \operatorname{tr}\!\left(
(\mathbf{A}(\mathbf{t})^\top \mathbf{A}(\mathbf{t}))^{-1}
\mathbf{A}(\mathbf{t})^\top \mathbf{A}(\mathbf{t})
(\mathbf{A}(\mathbf{t})^\top \mathbf{A}(\mathbf{t}))^{-1}
\right)\\
&= \sigma^2\,\operatorname{tr}\!\left(
(\mathbf{A}(\mathbf{t})^\top \mathbf{A}(\mathbf{t}))^{-1}
\right) =\sigma^2\,\frac{s_2(\mathbf{t})+m}{m s_2(\mathbf{t}) - (s_1(\mathbf{t}))^2},
\end{align*}
where $s_1(\mathbf{t})=\sum_{j=1}^m t_j$ and $s_2(\mathbf{t})=\sum_{j=1}^m t_j^2$. 
Note, $\frac{s_2(\mathbf{t})+m}{m s_2(\mathbf{t}) - (s_1(\mathbf{t}))^2}>0$ is due to Cauchy Schwartz inequality.

\begin{figure}
    \centering
    \includegraphics[width=0.48\linewidth]{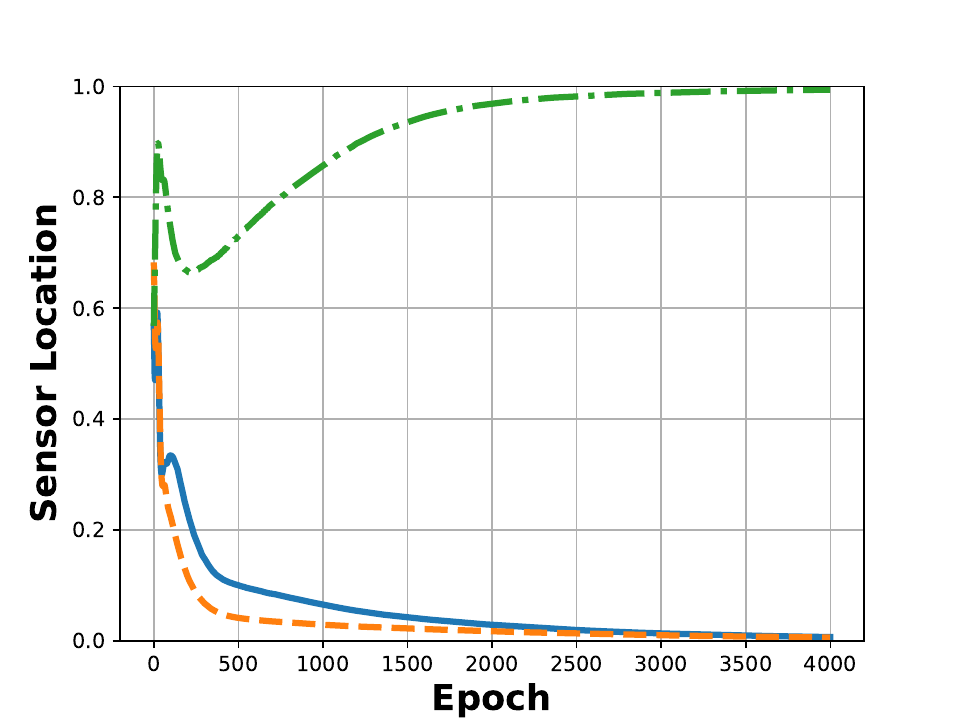}
    \includegraphics[width=0.48\linewidth]{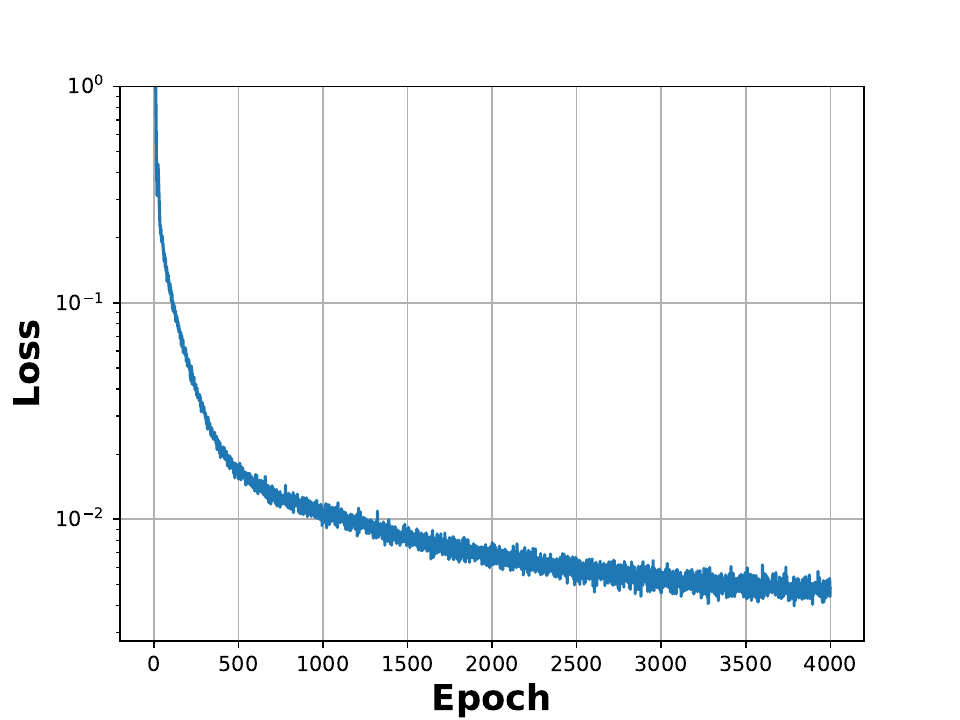}
    \caption{Evolution of sensor location and training loss during NODE training.  NODE is applied to \eqref{eq:exp_inverse_solution} for $m=3$ sensors. Prior distributions for model parameters are $a\sim\mathcal{U}(0.5,1.5)$ and $b\sim\mathcal{U}(1,2)$. Locations and model weights are trained simultaneously, using Adam optimizer, with learning rates of $10^{-1}$ and $10^{-3}$, respectively. Training lasts for \num{4000} epochs with batch size \num{1024}. Sensor locations stabilize to either~0 or~1 at the same time that the loss stabilizes.}
    \label{fig:locsExp}
\end{figure}

Consider now any arbitrary design points  $t_k,t_\ell\in(0,1)$. Then, replacing $(t_k,t_\ell)$ with 
\begin{equation}
    (\tilde t_k, \tilde  t_\ell) = \begin{cases} (0, t_k+ t_\ell)   & \text{if } t_k+ t_\ell<1 \\ 
                                                (1-t_k-t_\ell, 1) & \text{otherwise},

                                   \end{cases}
\end{equation}
reduces the overall objective function $F$, since $s_1$ remains unchanged and $s_2$ increases under this exchange. Hence $m-1$ design point must lie on the boundary. By fixing $m-1$ boundary points and viewing $F$ as a function of the remaining coordinate $t\in[0,1]$ shows that $F$ is strictly convex on $(0,1)$ and attains its minimum only at $t\in\{0,1\}$. Thus in any optimal design all points lie at the endpoints. An example showing the sensor locations evolving towards $0,1$ during training is shown in \Cref{fig:locsExp} for $m=3$ sensors.

Now let $k$ be the number of design points placed at $t=1$ (and $m-k$ at $t=0$). Then $s_1(\mathbf{t})=s_2(\mathbf{t})=k$, and the design criterion becomes
\[
F(k) = \sigma^2\,\frac{m+k}{k(m-k)},\qquad k\in\{1,\ldots,m-1\}.
\]

\begin{figure}
    \centering
    \includegraphics[width=0.75\linewidth]{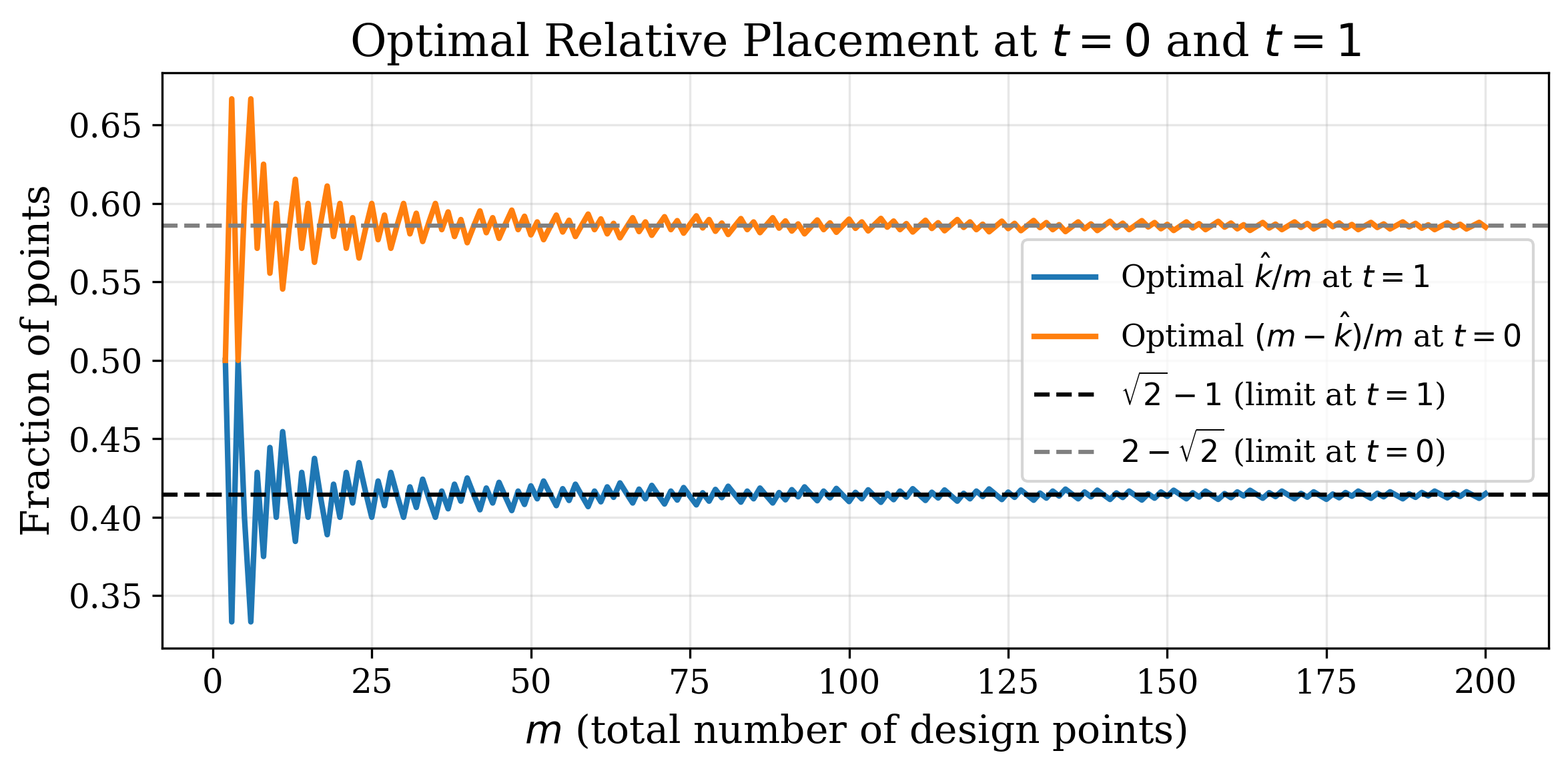}
\caption{
Optimal placement of design points at the boundary locations $t=0$ and $t=1$. Shown is the ratio of points assigned to each endpoint for increasing total design size $m$. The optimal design always places all sampling locations on the boundary, and the discrete optimal ratios converge as $m\to\infty$ to the continuous limits $\sqrt{2}-1 \approx 41.42\%$ at $t=1$ and  $2-\sqrt{2} \approx 58.58\%$ at $t=0$.
}
    \label{fig:optExp}
\end{figure}

Treating $k$ as a continuous variable in $(0,m)$, we see that $f$ is differentiable and strictly convex on $(0,m)$. Differentiation yields
\[
F'(k) = \sigma^2 \frac{k(m-k)-(m+k)(m-2k)}{k^2(m-k)^2} = \frac{k^2+2km - m^2}{k^2(m-k)^2}.
\]
Solving $F'(\hat k)=0$ gives the unique stationary point
\[
\hat k^2+2\hat km - m^2= 0 \Leftrightarrow \hat k= m(\sqrt{2} -1),
\]
which is therefore the global minimizer of the continuous relaxation. Consequently, for the discrete design problem $k\in\{1,\ldots,m-1\}$, the optimal integer choice is obtained by comparing the two nearest integers,
\[
k^- = \bigl\lfloor m(\sqrt{2}-1) \bigr\rfloor,
\qquad
k^+ = \bigl\lceil m(\sqrt{2}-1) \bigr\rceil,
\]
and selecting the one with the smaller value of $f(k)$, see \Cref{fig:optExp} 

Now that we have established the theoretical endpoint–optimal design, we investigate our NODE approach. For each total number of design points $m=2,\dots,200$ we apply \Cref{alg:node}, treating the $m$ locations $t_j\in[0,1]$ as trainable parameters that are updated jointly with a small reconstruction network. In each training iteration we draw a batch of parameter pairs $(a,b)$ with $a\sim\mathcal{U}(0.5,1.5)$ and $b\sim\mathcal{U}(1,2)$, and evaluate the forward model $y(t)=b\,\exp(a t)$ at the current design locations. Noise is introduced by adding i.i.d. Gaussian perturbations with standard deviation $\varepsilon = 0.05$ to the log-space transformed data, producing the noisy observations $\tilde y(t_j)$. The concatenated input vector $\tilde{\mathbf{y}}(\mathbf{t}) = [\tilde y(t_1),\dots,\tilde y(t_m)]$ is then formed and passed to a fully connected network with a single hidden layer of width~256 and ReLU activation. The network outputs estimates $(\hat a,\hat b)$ and is trained with a mean–squared error loss. The reconstruction weights and design locations are updated simultaneously using Adam with learning rates $10^{-3}$ for the reconstruction network and $10^{-1}$ for the design variables, a batch size of $1024$, and $10{,}000$ epochs per value of $m$. After each gradient step we project the learned locations back to the interval $[0,1]$.

\begin{figure}
    \centering
    \includegraphics[width=0.85\linewidth]{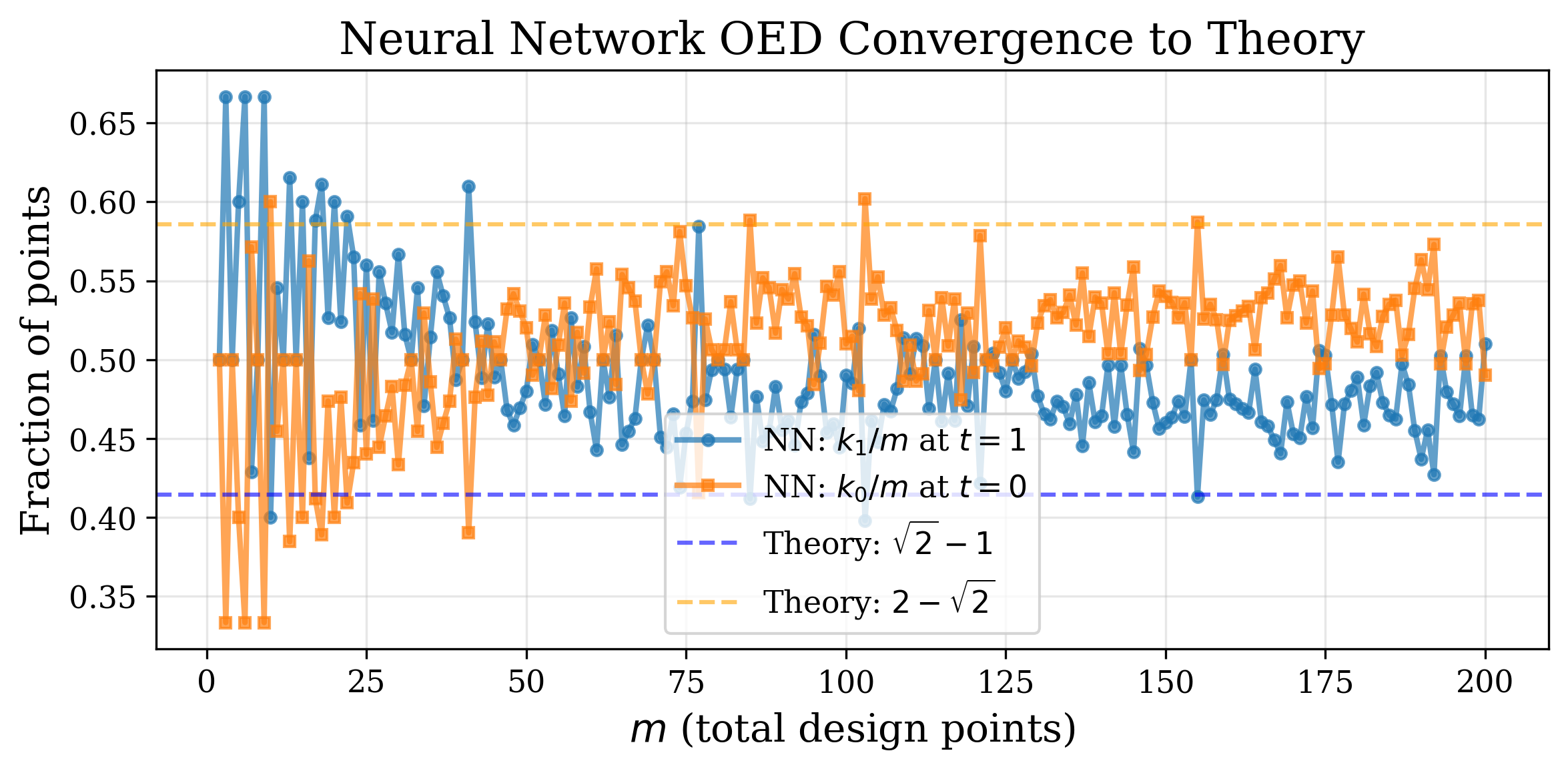}
    \caption{Neural OED convergence to the theoretical endpoint design for the scalar exponential model $y(t)=b\,\exp(a t)$.
    For each $m=2,\dots, 200$ we train the NODE algorithm (\Cref{alg:node}) with batch size $1024$, $10{,}000$ Adam steps (learning rates $10^{-3}$ for the reconstruction network and $10^{-1}$ for the design variables), Gaussian noise level $\varepsilon=0.05$ on the log-transformed data, and a single-hidden-layer ReLU MLP of width $256$ that reconstructs $(a,b)$ from noisy measurements and their locations. The plot shows, for each $m$, the fractions $k_1/m$ (blue, points near $t=1$) and $k_0/m$ (orange, points near $t=0$), together with the theoretical limits $\sqrt{2}-1\approx 0.4142$ and $2-\sqrt{2}\approx 0.5858$ (dashed lines). The learned designs systematically concentrate on the endpoints and the empirical fractions cluster around the theoretical values, demonstrating that the neural OED recovers both the qualitative endpoint structure and the optimal asymptotic splitting of design points.}
    \label{fig:appExp}
\end{figure}

For each value of $m$, the learned design is assessed by counting the number of points located near $t=0$ and near $t=1$.
We report on the count of design points at $0$ ($k_0)$ and $1$ ($k_1$) respectively. Specifically, we record the fractions $k_0/m$ and $k_1/m$. \Cref{fig:appExp} reports these fractions as functions of $m$ and compares them with the theoretical limits $\sqrt{2}-1\approx 0.4142$ (fraction of points at $t=1$) and $2-\sqrt{2}\approx 0.5858$ (fraction of points at $t=0$), compare also to \Cref{fig:optExp}. As designed, the NODE approach consistently drives the locations toward the endpoints of the interval, thereby recovering the qualitative endpoint–optimal structure predicted by the theory. Moreover, despite the deliberately simple network architecture, additive noise, and finite optimization budget, the learned fractions fluctuate around and increasingly concentrate near the theoretical limits as $m$ grows. The remaining scatter is consistent with finite-sample effects and imperfect optimization rather than a systematic deviation from the optimal design, underscoring that the neural approach is capable of reproducing the asymptotic optimal splitting of design points between $t=0$ and $t=1$.

\section{Numerics}\label{sec:numerics}
Here we present numerical results for applying NODE algorithm for sparse image reconstruction for \texttt{MNIST} grayscale images and X-ray computed tomography. 

\subsection{$\mathtt{MNIST}$ Image Reconstruction}\label{sec:mnist}

Which pixel locations in the \texttt{MNIST} images of handwritten digits are sufficient to accurately reconstruct the image or to reliably identify the digit it represents? We study this question next using our NODE method.

An image is an $m \times n$ array $\x$, where each $(i,j)$-entry corresponds to a pixel. The value $\x_{i,j} \in [0,1]$ represents the pixel intensity, ranging from black to white in grayscale images. Image reconstruction and deblurring~\cite{hansen2006deblurring} are classical problems in inverse problems. The aim here is to reconstruct an image $\x$ (assumed to be vectorized for notational convenience) from partial and noisy observations,
\begin{equation}\label{equ:image_model}
    \data = P_{\w}(\x) + \varepsilon.
\end{equation}
Here, the parameters correspond directly to pixel values, i.e., $\x = \y$. The observation $\data$ consists of a subset of pixel values selected by the masking operator $P_{\w}$, where $\w$ denotes the locations of the observed pixels, and is corrupted by additive noise $\varepsilon \sim \mathcal{N}(\mathbf{0}, \sigma^2 \mathbf{I})$.

As the problem is clearly ill-posed, we adopt a Bayesian regularization framework. We assume a prior distribution $\x \sim \pi_{\mathrm{pr}}$ on the image. Given an observation $\data$, the reconstruction is characterized by the posterior distribution $\pi_{\mathrm{post}}(\x \mid \data, \w)$,
\begin{equation}\label{equ:image_post}
    \pi_{\mathrm{post}}(\x \mid \data, \w)
    \propto
    \pi_{\mathrm{pr}}(\x)\,
    \pi_{\mathrm{like}}(\data \mid \x, \w),
    \qquad
    \pi_{\mathrm{like}}(\data \mid \x, \w)
    =
    \exp\!\left(
    -\frac{1}{2\sigma^2}
    \|P_{\w}(\x) - \data\|^2
    \right).
\end{equation}

The pixel locations $\w$ defining $P_{\w}$ constitute the experimental design. In inversion-oriented OED, the goal is to identify pixel locations that yield the most accurate reconstructions across the image distribution. In addition, we consider a goal-oriented setting in which full reconstruction is not the primary objective. Instead, the aim is to select pixel locations that enable accurate classification of each image according to the digit it represents.

We consider images from the $\mathtt{MNIST}$ data set~\cite{mnist,mnist-data}, which consists of \num{70000} grayscale images of handwritten digits. Each image is of size $28 \times 28$ and represents one of the ten digits $0$ through $9$. The image data are encoded as matrices $\x \in [0,1]^{28 \times 28}$, and each image is associated with a label vector $\boldsymbol{\ell} \in \{0,1\}^{10}$ identifying the corresponding digit. We treat the $\mathtt{MNIST}$ data set as a collection of \num{70000} independent samples drawn from the prior distribution $\pi_{\mathrm{pr}}$.

\paragraph{Optimality.}
We study optimal pixel selection under two inversion-oriented criteria, as well as a goal-oriented criterion. The first inversion-oriented approach uses a mean squared error (MSE) loss to select pixel locations that minimize the expected deviation from the posterior mean. This corresponds to the A-optimality design criterion and is defined by the loss
\begin{equation}\label{equ:mnist_mse}
    L_{\mathrm{MSE}}(\w)
    =
    \mathbb{E}_{\x,\varepsilon}
    \|\widehat{\x} - \x\|_2^2,
    \qquad
    \x \sim \pi_{\mathrm{pr}},\;
    \varepsilon \sim \mathcal{N}(\mathbf{0}, \sigma^2 \mathbf{I}),
\end{equation}
where the posterior mean $\widehat{\x}$ is conditioned on observations $\data = P_{\w}(\x) + \varepsilon$.

A second inversion-oriented criterion, reminiscent of E-optimality, focuses on minimizing the largest pointwise deviation from the posterior mean. The corresponding loss function is
\begin{equation}\label{equ:mnist_max}
    L_{\mathrm{max}}(\w)
    =
    \mathbb{E}_{\x,\varepsilon}
    \|\widehat{\x}(\w,\y) - \x\|_\infty^2
    =
    \mathbb{E}_{\x,\varepsilon}
    \Bigl(
    \max_{i,j}
    \bigl|\widehat{\x}_{i,j}(\w,\y) - \x_{i,j}\bigr|^2
    \Bigr),
\end{equation}
with $\x \sim \pi_{\mathrm{pr}}$ and $\varepsilon \sim \mathcal{N}(\mathbf{0}, \sigma^2 \mathbf{I})$.

In \Cref{equ:mnist_mse,equ:mnist_max}, the posterior mean $\widehat{\x}(\w,\y)$ may be viewed as a mapping from data to an estimator, $\y \mapsto \widehat{\x}(\w,\y)$. In the likelihood-free NODE setting, this mapping is approximated by a neural network $\Phi_\theta$. The reconstruction network $\Phi_\theta$ and the experimental design are learned jointly by minimizing empirical approximations of the loss functions.
\begin{equation}\label{equ:mnist_mse_nn}
    L_{\mathrm{MSE}}(\w,\theta)
    =
    \mathbb{E}_{\x,\varepsilon}
    \|\Phi_\theta(P_{\w}\x + \varepsilon) - \x\|_2^2\approx \tfrac{1}{K}\sum_{k=1}^K
    \|\Phi_\theta(P_{\w}\x_k+\varepsilon_k) - \x_k\|_2^2.
\end{equation}
Similarly, the max loss becomes
\begin{equation}\label{equ:mnist_max_nn}
    L_{\mathrm{max}}(\w,\theta)
    =
    \mathbb{E}_{\x,\varepsilon}
    \|\Phi_\theta(P_{\w}\x + \varepsilon) - \x\|_\infty^2\approx  \tfrac{1}{K}\sum_{k=1}^K
    \|\Phi_\theta(P_{\w}\x_k+\varepsilon_k) - \x_k\|_\infty^2.
\end{equation}

In the goal-oriented setting, the loss function is based on the categorical cross-entropy~\cite{crossentropy}. Here, the neural network $\Phi_\theta$ does not reconstruct the image $\x$, but instead predicts its label vector $\boldsymbol{v}$. The corresponding loss is
\begin{equation}\label{equ:mnist_class_nn}
    L_{\mathrm{CCE}}(\w,\theta)
    =
    -\mathbb{E}_{(\x,\boldsymbol{v}),\varepsilon}
    \bigl(
    \boldsymbol{v}^\top \log \Phi_\theta(P_{\w}\x + \varepsilon)
    \bigr)\approx -\tfrac{1}{K}\sum_{k=1}^K
    \boldsymbol{z}_k^\top
    \log\!\bigl(\Phi_\theta(P_{\w}\x_k+\varepsilon_k)\bigr).
\end{equation}
The output of $\Phi_\theta$ is a probability vector, where each entry represents the probability that the image corresponds to one of the ten digits. For all loss functions above, we jointly optimize over the observation locations $\w$ and the network parameters $\theta$ within our single-loop optimization framework \Cref{alg:node}.
\paragraph{Experimental settings.}
The \texttt{MNIST} data set is split into a training set of $K=\num{60000}$ images and a testing set of $\num{10000}$ images. Each image is corrupted with additive Gaussian noise of standard deviation $\sigma = 0.05$. 
  
All neural networks used in the experiments share the same multi-layer perceptron architecture, which depends only on the number of observation locations $|\w|=M$. Each network has $3M$ inputs: the $(i,j)$ pixel coordinates of the $M$ selected locations and the corresponding observed pixel intensities. These inputs are mapped to a single hidden layer with 512 neurons and ReLU activation. The output layer produces a reconstructed $28\times28$ image, yielding an output dimension of $784$. A sigmoid activation function is applied at the output to ensure that all reconstructed pixel values lie in the interval $[0,1]$.

Initial sensor placements are chosen as the $M$ pixel locations with the highest empirical variance across the training set, as discussed in \Cref{sec:finte-node}, while the network parameters $\theta$ are initialized randomly. All models are trained using the Adam optimizer~\cite{adam} with a learning rate of $10^{-3}$ for 50 epochs and a batch size of 64. Interpolation of the pixel location space is performed using bilinear interpolation~\cite{getreuer2011linear}.

For a fixed number of sensor locations, we compare the NODE designs against neural network models using non-optimized (fixed) designs. Specifically, we consider designs in which observation locations are selected either uniformly at random or as the highest-variance pixel locations used for NODE initialization. For these non-NODE baselines, only the network parameters $\theta$ are trained, while the observation locations $\w$ are held fixed throughout training. To assess the impact of the sensor budget, we perform reconstruction experiments with both NODE and non-NODE designs for varying numbers of allowed observation locations.

\paragraph{Results.} Our experiments feature sensor counts of $M=1,10,20,30,40,50,60,70,80,90,100$. For each $M$ we train models using the reconstruction loss functions~\eqref{equ:mnist_mse_nn} and~\eqref{equ:mnist_max_nn}, and the goal-oriented loss~\eqref{equ:mnist_class_nn}. 

We train models that use NODE designs, high variance designs, and random designs. For random designs, we repeat training twenty times for each experiment, using different random locations. 

\begin{figure}[h!!]
    \centering
    \includegraphics[width=1\linewidth]{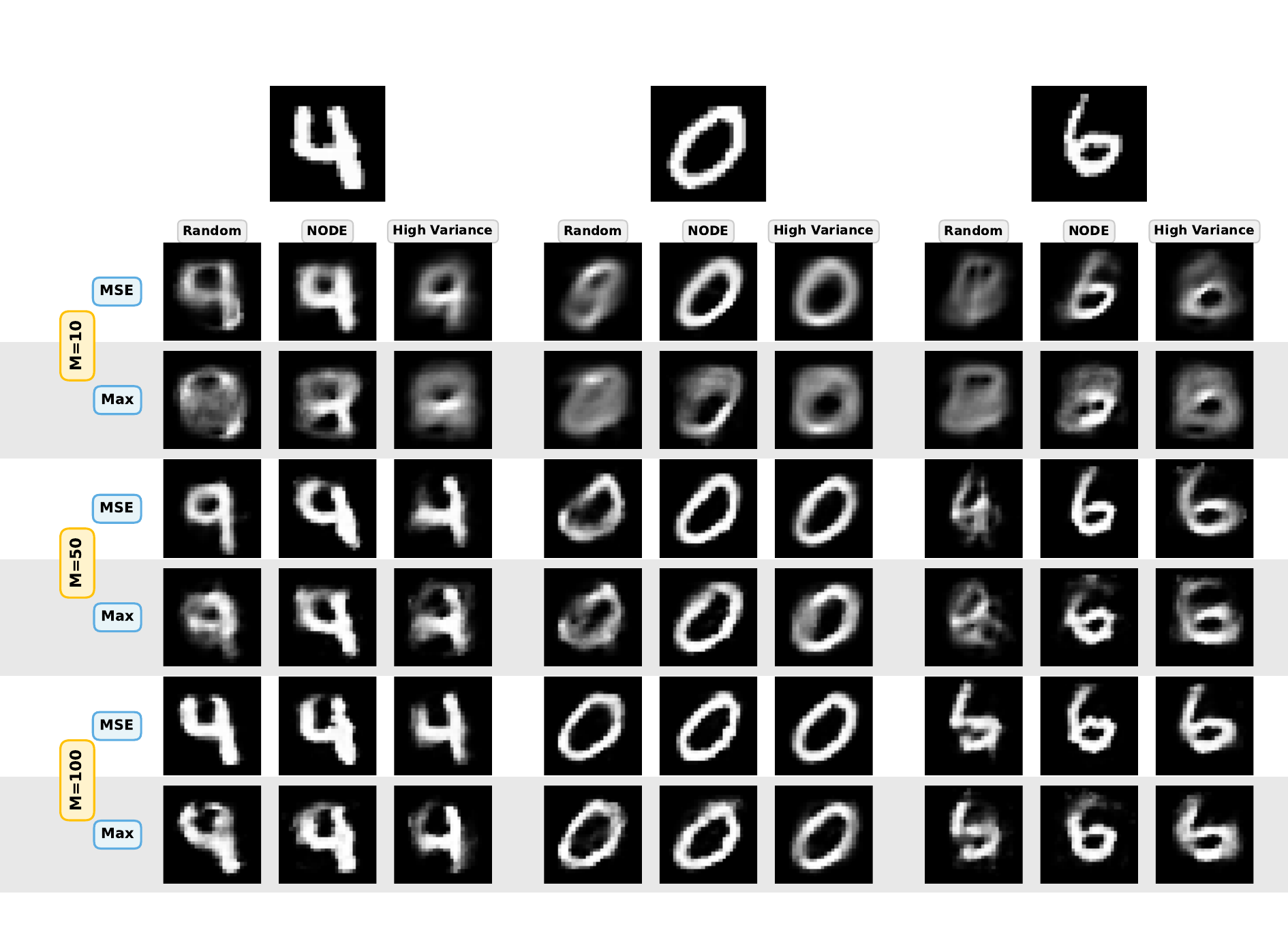}
    \caption{Sample image reconstructions of images, representing the digits 4,0,6. The top row displays the true images. Below, reconstructions are shown for different models. Models use NODE designs, random designs, or high variance designs, are trained using either the MSE loss~\eqref{equ:mnist_mse_nn} or max loss~\eqref{equ:mnist_max_nn}, and their designs use $M=10,50,100$ sensors.}
    \label{fig:digits}
\end{figure}

In \Cref{fig:digits}, we compare reconstructions of some example images from the testing set, from different reconstruction models. The models use designs for $M=10,50,100$ sensor locations.
The quality of reconstruction depends on which design is used and on the loss function used. The reconstructions representing random designs come from one of the randomly created designs and do not represent an average over the twenty random designs.

\begin{figure}[h!!]
    \centering
     \includegraphics[width=1\linewidth]{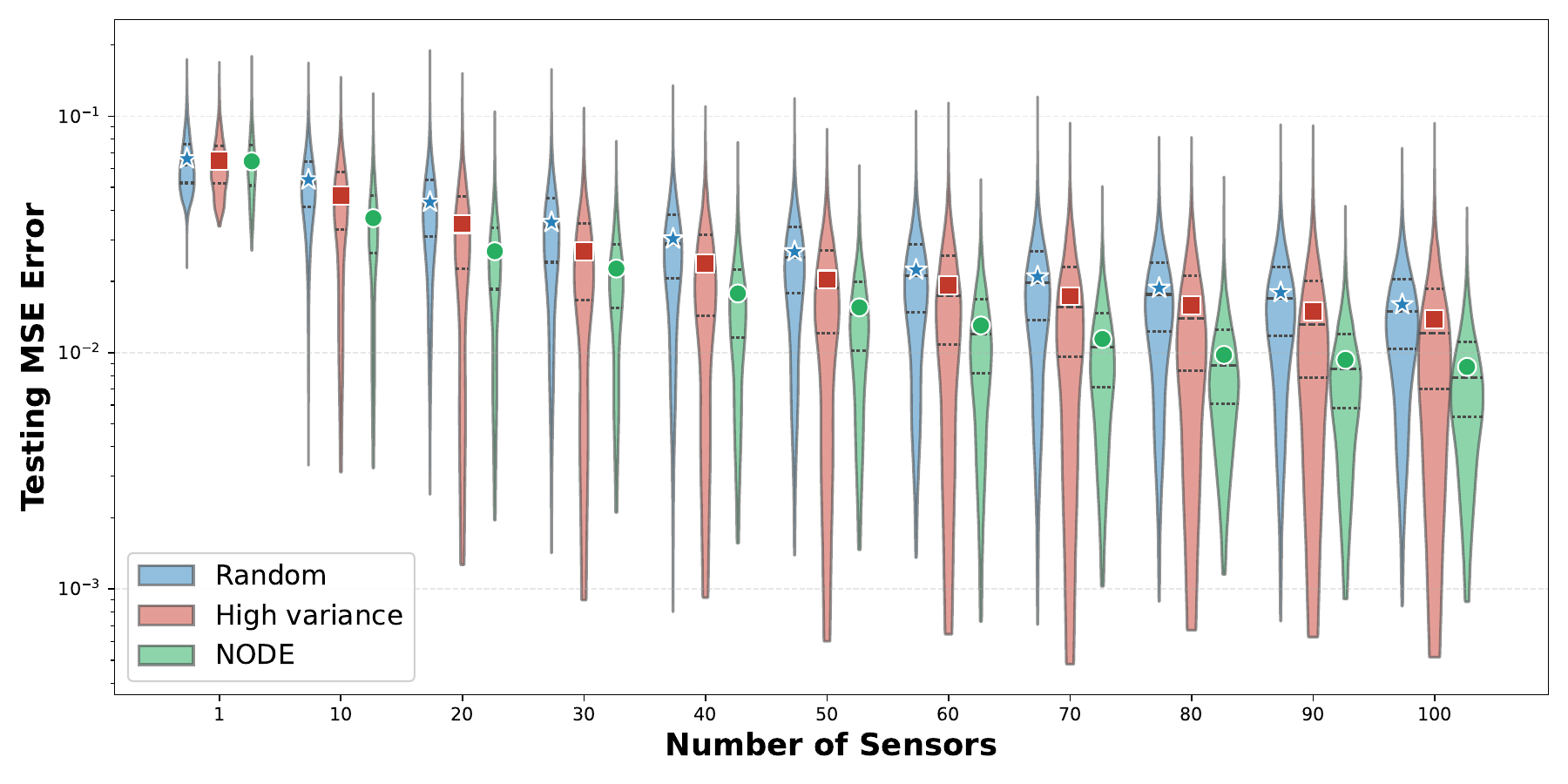}
      \includegraphics[width=1\linewidth]{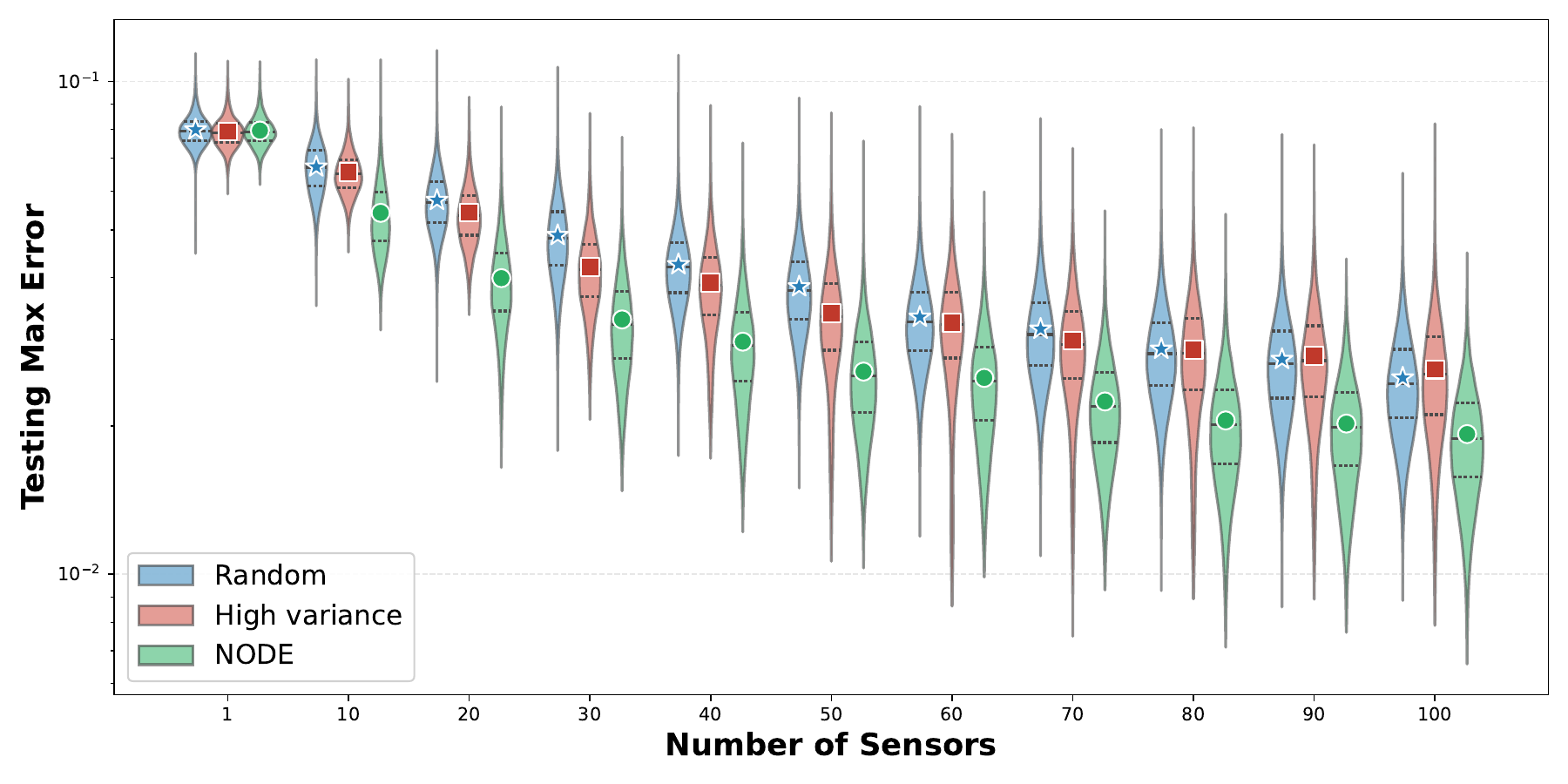}
    \caption{Log-scale comparison of the distribution of reconstruction error over the testing set of \num{10000} images.  Top figure shows reconstruction errors under the MSE loss~\eqref{equ:mnist_mse_nn}. Bottom figure shows reconstructions errors under the max loss~\eqref{equ:mnist_max_nn}. Distributions are computed different neural network architectures making use of $M=1,10,20,30,40,50,60,70,80,90,100$ sensor locations. Green distributions correspond to designs where locations are optimized by NODE. Blue distributions correspond to designs where locations fixed at randomly chosen values. Red distributions correspond to designs that use locations with the highest variance across the training set.}
    \label{fig:minst_violin}
\end{figure}

\begin{figure}[h!!]
    \centering
     \includegraphics[width=1\linewidth]{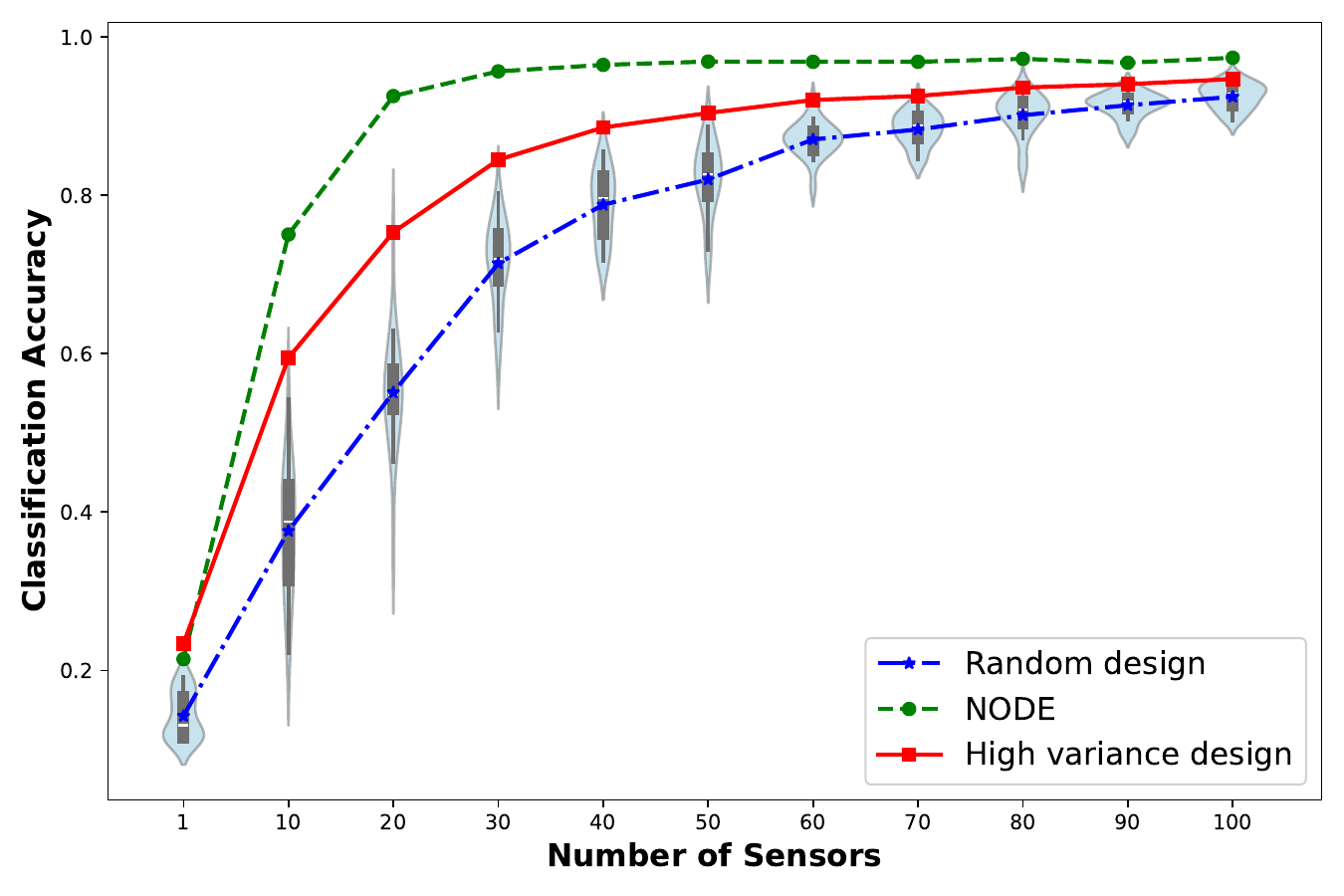}
    \caption{Comparison of the of classification accuracy on the testing set of \num{10000} images. Classification models are computed via neural network architectures making use of $M=1,10,20,30,40,50,60,70,80,90,100$ observable locations. The green plot correspond to NODE designs. The blue distributions correspond to designs where locations fixed at randomly chosen values. Red plot corresponds to designs that use locations with the highest variance.}
    \label{fig:minst_classify}
\end{figure}

\begin{figure}[h!!]
    \centering
     \includegraphics[width=1\linewidth]{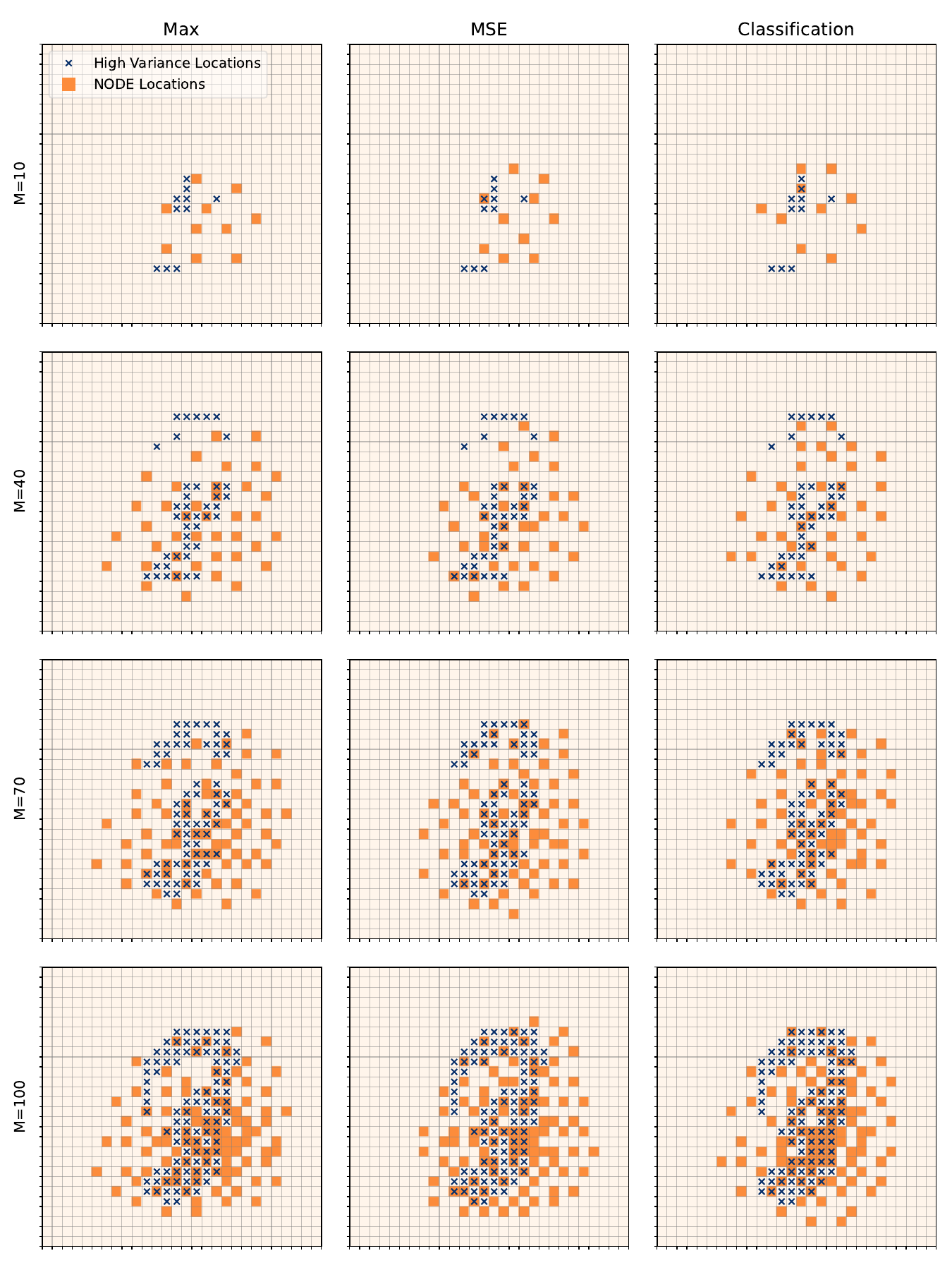}
    \caption{NODE design locations, on $28\times 28$ pixel grid, for $M=10,40,70,100$ sensors. Initial designs, the $M$ highest variance locations across the training set, are shown as a blue `x'. Initial locations are the same for all loss functions. Locations optimized by NODE are shown as orange squares.}
    \label{fig:minst_locs}
\end{figure}

The visual reconstructions confirms that the NODE designs yield better reconstructions when using either the MSE loss or the max loss. 
Checking the reconstruction errors, in \Cref{fig:minst_violin}, of the models on the set of test images confirms that NODE designs yield better reconstructions on average.
The distributions representing random designs describe errors across distributed across the testing set and all twenty random designs. The test set error distributions gives empirical evidence that NODE designs result, on average, in better reconstructions and in lower variance in the reconstruction error.
The goal-oriented setting does not involve reconstructions. Instead we gauge accuracy based on whether the model correctly classifies a digit. The classification accuracy is the percentage of images in the testing set that the model classifies correctly.
In \Cref{fig:minst_classify}, NODE designs result in models with better classification accuracy. Note that, for random designs, we have a distribution of classification accuracy over the twenty random designs.   
There is a small probability that random designs may outperform high variance designs. However, random designs are extremely unlikely to outperform NODE designs.
We lastly see how NODE designs compare to the high variance initial locations. In \Cref{fig:minst_locs}, NODE results in similar, but different, designs depending on the loss function used. 
The optimal locations cluster near the center pixels. This is to be expected since the digit strokes are in the centers of the images while the image borders  are black.

\subsection{Neural Optimal Design for X-ray Computed Tomography (CT)}\label{sec:ct}

Computed tomography (CT) is a well-established method for reconstructing an image of the cross-section of an object using projection data that represents the intensity loss or attenuation of a beam of X-rays as they pass through the object. According to the Lambert-Beer law \cite{hansen2021computed}, the attenuation of X-rays can be mathematically modeled as the line integrals or projection.

A simple mathematical model for X-ray computed tomography is given by the \emph{Radon transform}, which serves as the forward operator $f$ in \eqref{eq:forward-model}. In this setting, $f$ is a linear operator that maps a two-dimensional attenuation field $\x$ to its one-dimensional line integrals, commonly referred to as a \emph{sinogram}. The Radon transform can be expressed through the following integral equation:
\begin{equation} \label{eq:Radon}
    f [\x] (w, s) = \int_{\ell(w,s)} \x(r_1,r_2) \ \text{d} \ell = \int_{\mathbb R^2} \x(\boldsymbol{r}) \delta( s - \boldsymbol{r}\cdot \boldsymbol{n}(w) ) \ \text{d} \boldsymbol{r},
\end{equation}
where the middle formula represents a line integral along the line $\ell(w,s) = \{ \boldsymbol{r} |s-\boldsymbol{n}(w)\cdot\boldsymbol{r}\}$ with $\boldsymbol{r} = (r_1,r_2)\in \mathbb R^2$ representing Cartesian coordinates and $\boldsymbol{n}(w) = (\cos(w), \sin(w))$ the normal vector to $\ell$. A Radon transform often comprises $f [\x](w,s)$ where $w\in[0,2\pi)$ and $s\in[-1,1]$. We assume that $\x \equiv 0$ when $|r_1|+|r_2| > 1$ to ensure that the integrals above are bounded.

In this work we are interested in designing an optimal X-ray CT setup that utilizes a limited number of projections, i.e., we want to reconstruct $\x$ while optimally utilizing a few view angles $\w =\{w_1,\dots, w_m\}$. We refer to $f^{\w}$ as the projection operator associated to the angles in $\w$, i.e., the Radon transform \eqref{eq:Radon} subject to $w\in \w$. Under additive Gaussian noise, the inverse problem \eqref{eq:forward-model} reduces to
\begin{equation} \label{eq:ct-forward}
\y = f^{\w}[\x] + \varepsilon,
\end{equation}
where, with a slight abuse of notation, $\x \in \mathbb{R}^{n \times n}$ denotes the discrete unknown cross-sectional image, and $\varepsilon$ represents additive Gaussian measurement noise. The forward operator $f^{\w} : \mathbb{R}^{n \times n} \to \mathbb{R}^{\rho \times m}$ maps the image to its corresponding sinogram, where $\rho$ denotes the number of X-ray sensors, represented by a discretization of the interval $[-1,1]$ for the detector coordinate $s$. The noise is modeled as $\varepsilon \sim \mathcal{N}(\mathbf{0}, \sigma^2 \mathbf{I}_{\rho \times m})$, with $\sigma$ denoting the noise standard deviation. The optimal design, for the X-ray CT problem comprises finding the best projection angles $\w$ to reconstruct cross-sectional image $\x$ that solves the inverse problem \eqref{eq:ct-forward}.

\paragraph{NODE Formulation of X-ray CT.}

In this section we reformulate the X-ray CT inverse problem into a join design and reconstruction problem described in \Cref{sec:method}.

We start with a dataset $\{\x_k\}_{k=1}^K$ of ground truth cross-sectional images. In this work we use the Low Dose Parallel Beam (LoDoPaB) dataset \cite{Leuschner2021-kg,Leuschner2019-dd} which comprises over \num{40000} cross-sectional thorax images of around 800 patients. This dataset is split over the training data containing around \num{36000} slices and a validation dataset with around \num{4000} cross-sectional images.

Following \eqref{eq:deep-oed2} we use the NODE loss function for the X-ray CT problem as
\begin{equation} \label{eq:loss-CT}
    \underset{\theta, \w}{\text{arg min}} \  \mathbb E_{\alpha,\varepsilon} \| \Phi_\theta ( \mathcal I ( f^{\w  }[\x] + \varepsilon ) ) - \x \|_2^2.
\end{equation}

Here, $f^{\w  }$ is the sparse angle Radon transform introduced above, and $\Phi_\theta$ is a parameterized family of inversion method, and $\mathcal I:\mathbb R^{\rho\times m} \to \mathbb R^{n\times n}$ is an interpolation mapping, as described in \Cref{sec:finte-node}.

In this work, we choose $\Phi_\theta$ to be a convolutional U-Net~\cite{ronneberger2015u}. This architecture follows an encoder–decoder structure: the contracting path captures multiscale features through successive convolution and pooling operations, while the expanding path enables precise localization via upsampling and concatenation with corresponding feature maps from the encoder. We adopt this architecture because U-Nets are particularly effective at extracting and integrating localized features, making them well suited for processing sinogram data.

U-Nets typically require the input and output to have identical spatial dimensions. To satisfy this requirement, we introduce an interpolation operator $\mathcal{I}$ that resamples the sinogram so that its dimensions match those of the target image.

Solving the optimization problem in \eqref{eq:loss-CT}, for example using gradient descent methods, requires knowledge of the partial derivatives with respect to both $\theta$ and $\w$. While computing derivatives with respect to $\theta$ is standard practice, commonly referred to as back-propagation, we also require the partial derivatives of $f^{\w} [\x]$, with respect to $\w$. We can differentiate \eqref{eq:Radon} with respect to one projection angle $w$ to obtain
\begin{equation}
    \begin{aligned}
        \partial_{w} f^{w}[\x](w,s) &= \int_{\mathbb R^2} \x( \boldsymbol{r} ) \partial_{w} \delta(s - \boldsymbol{r}\cdot \boldsymbol{n}(w)) \ \text{d} \boldsymbol{r} 
        = -\int_{\mathbb R^2} \x(\boldsymbol{r}) (\boldsymbol{r}\cdot \boldsymbol{t}(w) ) \delta'(s - \boldsymbol{r}\cdot \boldsymbol{n}(w)) \ \text{d} \boldsymbol{r} \\
        &= -\int_{\mathbb R^2} \x(\boldsymbol{r}) (\boldsymbol{r}\cdot \boldsymbol{t}(w) ) \partial_s \delta(s - \boldsymbol{r}\cdot \boldsymbol{n}(w)) \ \text{d} \boldsymbol{r} 
        = -\partial_s \int_{\mathbb R^2} \x(\boldsymbol{r}) (\boldsymbol{r}\cdot \boldsymbol{t}(w) ) \delta(s - \boldsymbol{r}\cdot \boldsymbol{n}(w)) \ \text{d} \boldsymbol{r} \\
        &= - \partial_{s} f^w[\x(\boldsymbol{r}\cdot \boldsymbol{t}(w) )](w,s).
    \end{aligned}
\end{equation}
Here, we used the fact $\boldsymbol{t}(w) := \boldsymbol{n}'(w) = (-\sin(w), \cos(w))$, i.e., the derivative of a normal vector is the tangential vector. However, in practice we utilize auto-differentiation tools, e.g., autograd feature from PyTorch, to calculate these partial derivatives.

To solve the minimization problem in \eqref{eq:loss-CT}, we employ the Adam optimization algorithm \cite{adam} with a mini-batch strategy using a batch size of 128 of the training LoDoPaB dataset. The learning rate is fixed at $2\times 10^{-3}$, and the optimization is performed for a total of 150 epochs.

We evaluate the performance of the proposed method on the LoDoPaB validation dataset. Specifically, we solve the neural optimal design problem \eqref{eq:loss-CT} for two cases: one where the size of $\w$ is $45$, and another where it is $10$. We remind the reader that a typical CT reconstruction involves $\mathcal{O}(100)$ projection angles; therefore, both cases considered in this experiment correspond to highly challenging X-ray CT reconstruction problems.

As the initial condition for the optimization problem \eqref{eq:loss-CT}, we initialize all projection angles to be uniformly distributed within the interval $[0, \tau)$, where $\tau = \pi$ when $|\w| = 10$ and $\tau = \pi/6$ when $|\w| = 45$. These differing initializations are chosen to highlight the adaptivity of the proposed method. Following the training phase, the projection angles are optimized to their final configurations.

\begin{figure}
    \centering
    \includegraphics[width=\linewidth]{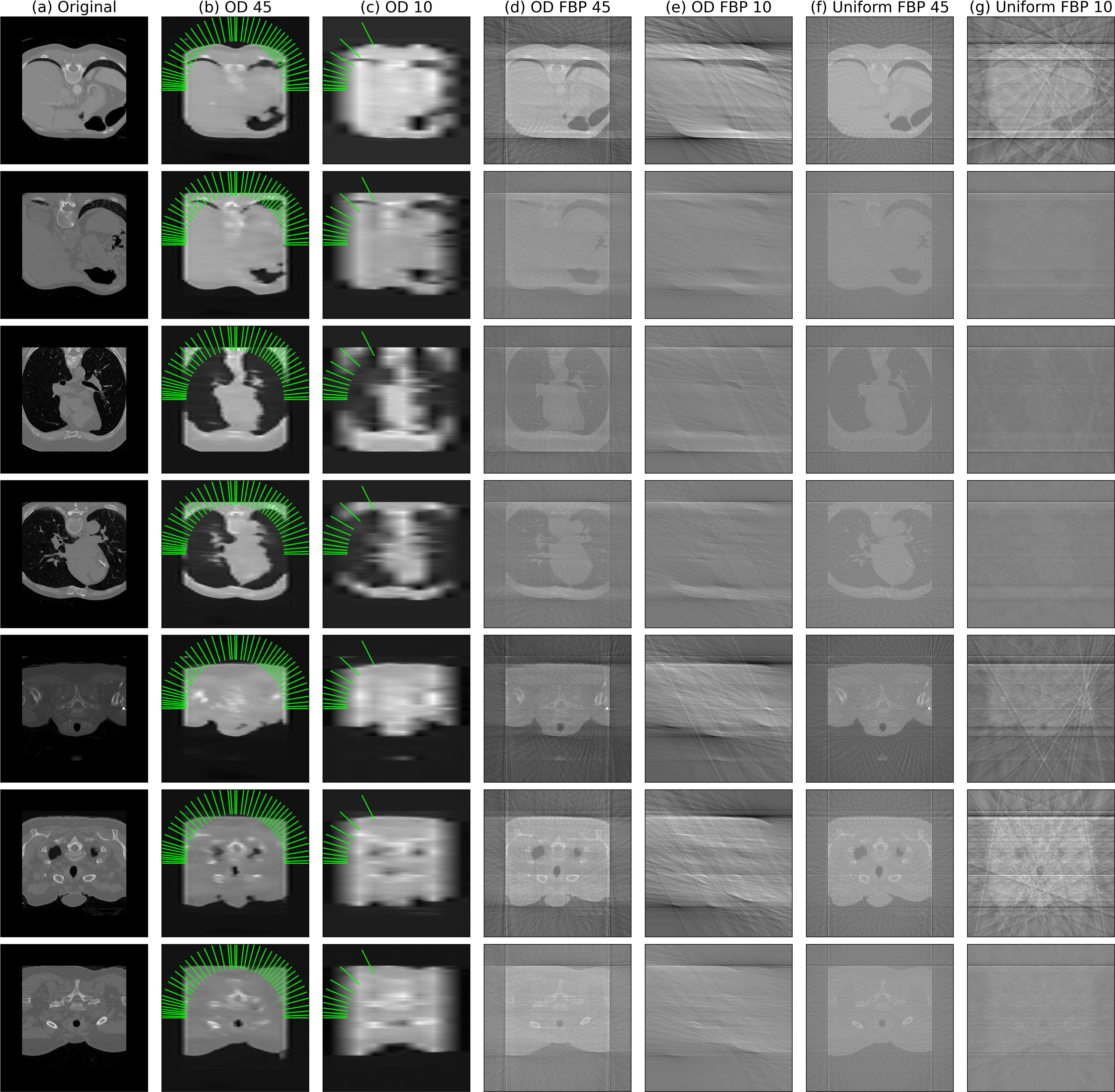}
    \caption{NODE suggested of X-ray projection angles and the corresponding reconstruction. NODE angles are indicated with green colors. (a) original cross-sectional thorax image (b) NODE of 45 projection angles and reconstruction (c) NODE of 10 projection angles and reconstruction (d) FBP reconstruction with NODE 45 design angles (e) FBP reconstruction with NODE 10 design angles (f) FBP reconstruction with 45 equidistant design angles (g) FBP reconstruction with 10 equidistant design angles.}
    \label{fig:xray-plots}
\end{figure}

We present the NODE-designed projection angles in \Cref{fig:xray-plots}. Column (a) shows representative cross-sectional images from the LoDoPaB test dataset. The NODE-optimal projection angles are shown in columns (b) and (c) for designs with 45 and 10 projection angles, respectively. In both cases, the learned designs differ markedly from the uniformly spaced angles commonly used in sparse-view X-ray CT. In particular, NODE allocates more view angles near the horizontal and vertical directions. Some reconstruction artifacts are visible, especially for the extremely sparse case of 10 projections; these could likely be mitigated by employing a more specialized reconstruction architecture.

We further compare filtered backprojection (FBP) reconstructions obtained using NODE-designed angles and uniformly spaced angles. Columns (d) and (e) in \Cref{fig:xray-plots} show FBP reconstructions with NODE angles, while columns (f) and (g) correspond to equidistant angle placement. The NODE-based designs yield noticeably improved image quality, and FBP reconstructions using NODE angles better preserve structural features and discontinuities than those obtained with uniform sampling.

\paragraph{Adaptive NODE for X-ray CT.}
Next, we modify the NODE cost function to enable adaptive updates of the projection angles. Following the methodology described in \Cref{sec:adaptive-NODE}, we implement an adaptive experimental design framework for X-ray computed tomography. Although, in practice, measurements are acquired from patients in a clinical setting, in this work we instead generate measurements from cross-sectional images in the LoDoPaB test dataset using a simulated Radon transform.

to summarize, suppose that we have a true image $\x^{\text{true}}$ but unknown to the method. In the $i$th ($i\geq 0$) step of the adaptive NODE, we suppose that we have $\y^{i-1}\in \mathbb R^{\rho\times n}$ as sinogram already collected from $\x^{\text{true}}$ which correspond to view angles $\w^{i-1} = \{ w_1,\dots,w_n \}$. We now look for new set of angles $\w^{i} = \{ w_{n+1},\dots, w_{n+p} \}$ to complement $\w^{i-1}$ in the sense

\begin{equation} \label{eq:incremental-NOD}
    \underset{\theta, \w}{\text{arg min}}~ \mathbb E_{\x,\varepsilon} \ \left\| \Phi_{\theta } \left( \mathcal I^i \left( 
    \begin{bmatrix}
    & f^{ \w  }[\x] + \varepsilon \\
    & \y^{i-1}
    \end{bmatrix}
    \right) \right) - \x \right\|_2^2 .
\end{equation}
As discussed in \Cref{alg:adaptive-node} each new iteration of the adaptive NODE requires a new reconstruction architecture $\Phi_{\theta}^{i}$, e.g., number of cells and layers in the inversion network. However, if we assume that we are in a sparse angle imaging setup, i.e., the number of view angles are significantly smaller than image dimensions, we can fix the architecture of $\Phi_{\theta}$ and only modify the interpolation operator $\mathcal I^{i}$ to map data into image dimensions.

In this example, we choose the design of the inversion network $\Phi_\theta$ to be identical to previous section and choose increments of $5$ view angles, i.e., $| \w^{i}| = 5$ for all $i\geq 0$. The method is initiated $(i=0)$ by choosing $\w_0=\{ 0+n\pi/5 \mid n=1,\dots,5 \}$. For the rest of iterations, we initiate optimization \eqref{eq:incremental-NOD} by choosing $\w_i=\{ 0+n\pi/30 \mid n=1,\dots,5 \}$ and create $\y^{i}$ by concatenating previous sinograms with the new ones, i.e.,
\begin{equation}
    \y^{i} = ( \y^{i-1}, f^{\w^\star }[\x^{\text{true}}] + \varepsilon ), \qquad i\geq 1,
\end{equation}
where $\w^{\star}$ is the approximate solution of \eqref{eq:incremental-NOD} in the $i$th iteration, and $\varepsilon$ contains a Gaussian noise $\varepsilon\sim \mathcal N(\mathbf{0},\sigma^2\mathbf{I})$, with noise level to $\sigma = 0.01\times \| f^{\w^\star }[\x^{\text{true}}]\|_2$.  We perform the incremental NODE iterations for $i=0,1,2$.

We choose a small incremental addition of projection angles to better showcase the potential of NODE in challenging inverse problems. Moreover, visualizing and distinguishing the impact of different projection angles is inherently difficult when many angles are involved; limiting the number of added projections allows reconstruction improvements and differences between projection angle selections to be more clearly observed.

\begin{figure}
    \centering
    \includegraphics[width=0.75\linewidth]{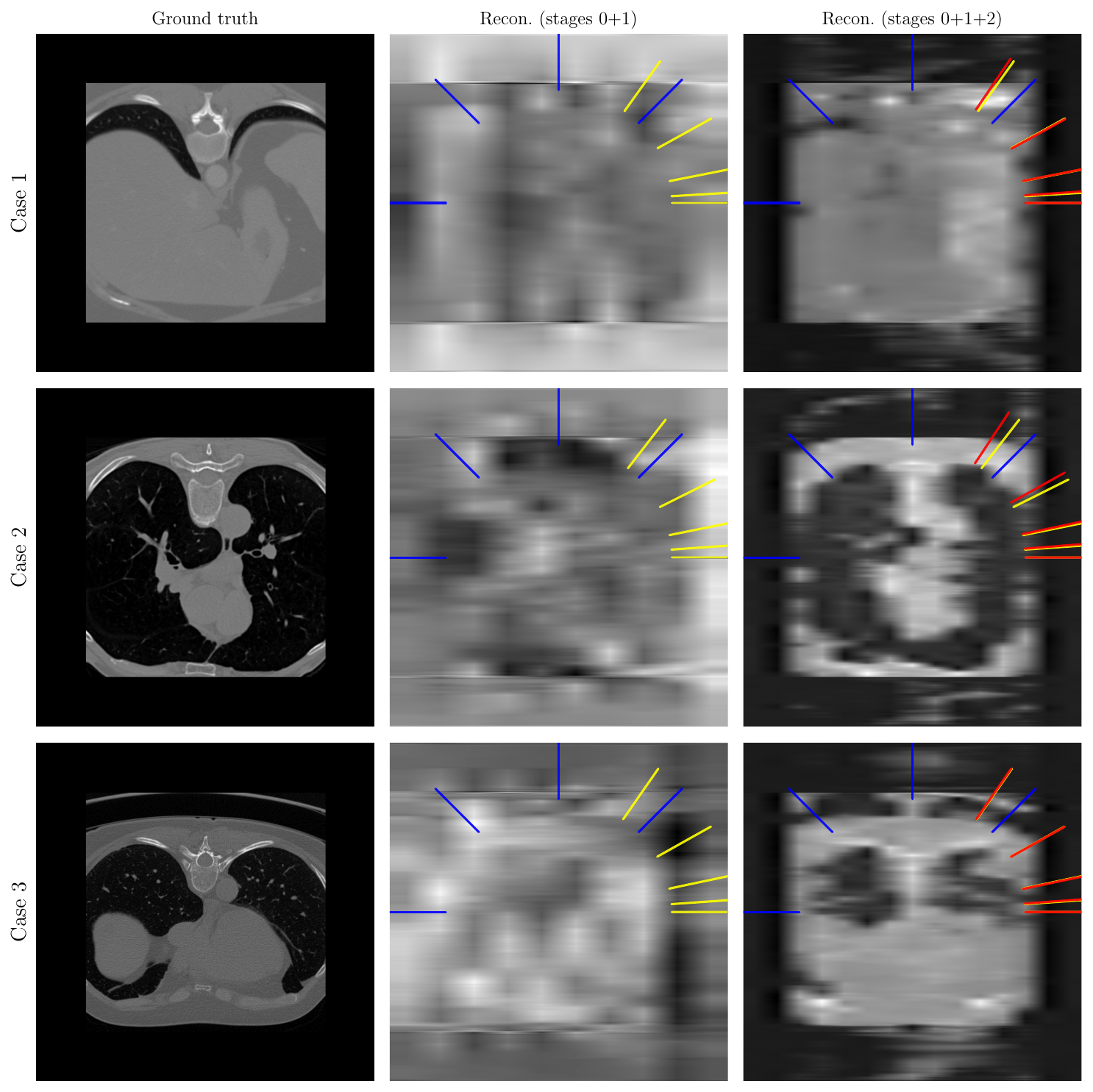}
    \caption{the results for the adaptive NODE method for three test images. The first column shows the ground truth images where the second and third columns corresponds to the first and second iterations of the adaptive NODE method. The blue lines indicate the initial uniformly separated angles, while the yellow and red corresponds to NODE angles, after the first and second iterations, respectively.}
    \label{fig:incremental-NOD}
\end{figure}

We apply the incremental NODE framework to three test images from the LoDoPaB test dataset, and the results are presented in \Cref{fig:incremental-NOD}. Each row corresponds to a single test case, with the first column showing the ground-truth image. The directions of the projection angles are indicated by colored lines: the blue lines denote the initial projection angles, chosen uniformly; the yellow lines in the middle column show the NODE-selected projection angles after the first adaptive iteration; and the red lines in the right column indicate the projection angles selected in the second iteration. The results demonstrate that successive incremental NODE iterations lead to improved reconstructions while yielding case-specific and adaptive view-angle selections.

\section{Conclusion and Outlook}\label{sec:conclusion}

We introduced \emph{Neural Optimal Design of Experiments} (NODE), a learning-based framework for optimal experimental design that circumvents the classical bi-level optimization paradigm. Building on deep OED~\cite{siddiqui2024deep}, NODE replaces sparsity-inducing weight optimization with direct optimization over measurement locations themselves. By leveraging smooth interpolation of the design space, this approach significantly reduces problem dimensionality while avoiding $\ell^1$ regularization and associated hyperparameter tuning. The resulting method is flexible, scalable, and straightforward to implement.

For a scalar exponential growth model, we derived a benchmark problem with a closed-form A-optimal design and proved that all optimal designs concentrate at the boundary of the experimental interval, with a precise asymptotic splitting between endpoints. NODE consistently recovers this structure across a wide range of design budgets, demonstrating that joint neural training can reproduce classical optimal design behavior. In higher-dimensional settings, NODE designs outperform random and variance-based baselines, yielding improved reconstruction and classification accuracy. In sparse-view CT experiments, the learned projection angles deviate substantially from uniform sampling and align with known structural preferences, indicating that NODE captures meaningful geometric information intrinsic to the inverse problem.

One limitation of the current NODE framework is that the architecture of the reconstruction network $\Phi_\theta$ depends on the sensor budget $M$. As a result, identifying optimal designs for different budgets requires retraining the model, which can be computationally expensive when studying budget–performance trade-offs. A promising direction to address this issue is the use of autoencoders~\cite{Afkham2024,Chung_autoencoder2024,Chung_autoencoder2025} to embed the output space into a lower-dimensional latent representation. Applying NODE in latent space could enable more compact architectures and amortize training across multiple sensor budgets.

The interpolation strategy employed for finite design spaces draws on techniques from image processing. While effective in practice, such methods are not intrinsic to optimal experimental design. Developing interpolation schemes specifically tailored to discrete design optimization is an important avenue for future work. Optimization over interpolated discrete spaces can be interpreted as a finite-dimensional problem with infinitely many constraints, linking NODE to semi-infinite programming~\cite{semi-inf}. Mollifier-based approaches~\cite{mollifier,tadmor2002}, which concentrate information locally around admissible locations while smoothly decaying elsewhere, may offer a principled alternative.

Finally, neither NODE nor existing single-level learning-based OED methods~\cite{siddiqui2024deep} directly address D-optimality. Extending NODE to this setting would require access to the full posterior distribution rather than point estimates. Generative models such as diffusion models or conditional normalizing flows offer a natural path forward by enabling posterior sampling within the loss function. In a single-level formulation, both the generative model and the experimental design could be trained jointly. This has already received significant attention in recent work~\cite{Kennamer_Walton_Ihler_2023,Dong2025,orozco2024probabilistic}. Alternatively, a two-stage strategy may be more appropriate in real-world applications, where a data-driven generative model trained on simulated data is combined with simulation-based inference~\cite{rosso2025} using experimental observations to optimize the design.

\bmhead{Acknowledgments} This work is partially supported by Research Council of Finland under grant numbers 359186 and 371523 (Babak Maboudi Afkham) and by the National Science Foundation (NSF) under grand number 2038118 (John E. Darges).

\bibliography{refs}

\end{document}